\documentclass{article} % For LaTeX2e
\usepackage{iclr2026,times}

\usepackage{amsmath,amsfonts,bm}

\def\eqref#1{equation~\ref{#1}}
\def\1{\bm{1}}

\DeclareMathAlphabet{\mathsfit}{\encodingdefault}{\sfdefault}{m}{sl}
\SetMathAlphabet{\mathsfit}{bold}{\encodingdefault}{\sfdefault}{bx}{n}

\usepackage{url}
\usepackage[utf8]{inputenc} % allow utf-8 input
\usepackage[T1]{fontenc}    % use 8-bit T1 fonts
\usepackage{booktabs}       % professional-quality tables
\usepackage{threeparttable}
\usepackage{amsfonts}       % blackboard math symbols
\usepackage{nicefrac}       % compact symbols for 1/2, etc.
\usepackage{microtype}      % microtypography
\usepackage[svgnames]{xcolor}
\usepackage{soul}      % 核心高亮宏包（零冲突）

\definecolor{myhighlight}{RGB}{255,255,153}  % 浅黄色
\definecolor{deepblue}{RGB}{100,149,237}    % 钢蓝色

\sethlcolor{myhighlight}  % 设置默认高亮颜色
\usepackage{hyperref}
\usepackage{multirow}

\usepackage{graphicx}
\usepackage{amsmath,amssymb,amsthm}
\usepackage{tcolorbox}
\usepackage{enumitem}
\usepackage{subfigure}
\usepackage{array}
\usepackage{balance}
\usepackage{colortbl}
\usepackage{stfloats}
\usepackage{tabularx}
\usepackage{textcomp}
\usepackage{verbatim}
\usepackage{bm}
\usepackage{adjustbox}
\usepackage{float}
\usepackage{wrapfig}

\makeatletter
\newif\if@restonecol
\makeatother

\usepackage[linesnumbered,ruled,vlined]{algorithm2e}
\usepackage{algpseudocode}

\SetKwComment{Comment}{$\triangleright$\ }{}
\SetKwRepeat{Do}{do}{while}
\definecolor{b_proj}{HTML}{D71D3D}
\definecolor{b_mse}{HTML}{FFD966}
\definecolor{b_act}{HTML}{999999}

\renewcommand{\paragraph}[1]{\noindent\textbf{#1}}
\renewcommand{\subparagraph}[1]{\noindent\textbf{\underline{#1.}}}
\newcommand{\revise}[1]{{\color{black} #1}} % 

\title{Uni-X: Mitigating Modality Conflict with a Two-End-Separated Architecture for Unified Multimodal Models}

\author{Jitai Hao$^{1,*}$ \quad
  Hao Liu$^{2,*}$ \quad
  Xinyan Xiao$^{2}$ \quad
  Qiang Huang$^{1,\dagger}$ \quad
  Jun Yu$^{1,\dagger}$ \\ [6pt]
  $^1$School of Intelligence Science and Engineering, Harbin Institute of Technology (Shenzhen) \\
  $^2$Baidu Inc. \\ % \quad $^3$Pengcheng Laboratory \\ % 将较短的单位放在一行
  \texttt{jitaihao@outlook.com, \{huangqiang, yujun\}@hit.edu.cn} \\
  \texttt{\{liuhao24, xiaoxinyan\}@baidu.com} \\
  $^{*}$Equal contribution. \quad $^{\dagger}$Corresponding authors.
}

\iclrfinalcopy % Uncomment for camera-ready version, but NOT for submission.

\begin{document}
\maketitle

% 修复新版 fancyhdr 导致页眉文字丢失的 Bug
\ificlrfinal
    \fancyhead[L]{Published as a conference paper at ICLR 2026}
\else
    \fancyhead[L]{Under review as a conference paper at ICLR 2026}
\fi

\begin{abstract}
%%% background & motivations
Unified Multimodal Models (UMMs) built on shared autoregressive (AR) transformers are attractive for their architectural simplicity. However, we identify a critical limitation: when trained on multimodal inputs, modality-shared transformers suffer from severe gradient conflicts between vision and text, particularly in shallow and deep layers. We trace this issue to the fundamentally different low-level statistical properties of images and text, while noting that conflicts diminish in middle layers where representations become more abstract and semantically aligned. 
%%% architecture
To overcome this challenge, we propose Uni-X, a two-end-separated, middle-shared architecture. Uni-X dedicates its initial and final layers to modality-specific processing, while maintaining shared parameters in the middle layers for high-level semantic fusion. This X-shaped design not only eliminates gradient conflicts at both ends but also further alleviates residual conflicts in the shared layers.
%%% experiments
Extensive experiments validate the effectiveness of Uni-X. Under identical training conditions, Uni-X achieves superior training efficiency compared to strong baselines. When scaled to 3B parameters with larger training data, Uni-X matches or surpasses 7B AR-based UMMs, achieving a GenEval score of 82 for image generation alongside strong performance in text and vision understanding tasks.
These results establish Uni-X as a parameter-efficient and scalable foundation for future unified multimodal modeling.
Our code is available at \url{https://github.com/CURRENTF/Uni-X}.
\end{abstract}

\section{Introduction}
\label{sect:intro}

Vision-Language Models (VLMs) have demonstrated remarkable progress in multimodal understanding and reasoning, enabled by combining Large Language Models (LLMs) with powerful visual encoders~\citep{llava, liu2024llavanext, wang2024qwen2vlenhancingvisionlanguagemodels, team2024gemma2}. 
%%%
Motivated by this success, recent efforts to add image generation have led to \textbf{Unified Multimodal Models (UMMs)}~\citep{team_chameleon_2025, wang2024emu3, wu2024vila}.
%%%
However, many advanced UMMs rely on increasingly complex system designs to boost performance, including the addition of semantic image encoders~\citep{wu_janus_2024, chen_janus-pro_2025, deng_emerging_2025, wu_omnigen2_2025, lin2025toklip}, the hybridization of autoregressive and diffusion paradigms~\citep{wu_omnigen2_2025, zhao2024monoformer, ge2025seedxmultimodalmodelsunified, deng_emerging_2025, xie_show-o2_2025, transfusion}, or the introduction of task-specific branches and experts~\citep{deng_emerging_2025, liao_mogao_2025, li_unifork_2025}. 
%%%
While effective, this added complexity hinders scalability, limiting the degree of parameter sharing and reducing the potential for mutual benefits across tasks and modalities.

In contrast, \textbf{autoregressive (AR) UMMs} offer a simple yet powerful alternative.
%%%
By treating visual inputs as a ``foreign language'' through vector quantization (VQ)~\citep{vqvae_oord2018neuraldiscreterepresentationlearning, vqgan_esser2021tamingtransformershighresolutionimage}, they unify text and vision into a consistent token sequence, naturally extending the language-centric paradigm of LLMs~\citep{wu_liquid_2025, wang2024emu3}.
%%%
Despite this simplicity, our experiments reveal a fundamental challenge: \textbf{fully modality-shared transformers trained jointly on multimodal inputs exhibit severe gradient conflicts.} 
Originally studied in multi-task learning~\citep{grad_surgery, recon_grad}, we are the first to transfer this concept to UMMs, uncovering inter-modality conflicts that hinder convergence and performance.

\begin{wrapfigure}{r}{0.5\textwidth}
  \vspace{-2.0em}
  \begin{minipage}{\linewidth}
    \begin{figure}[H]
      \centering
      \includegraphics[width=0.95\columnwidth]{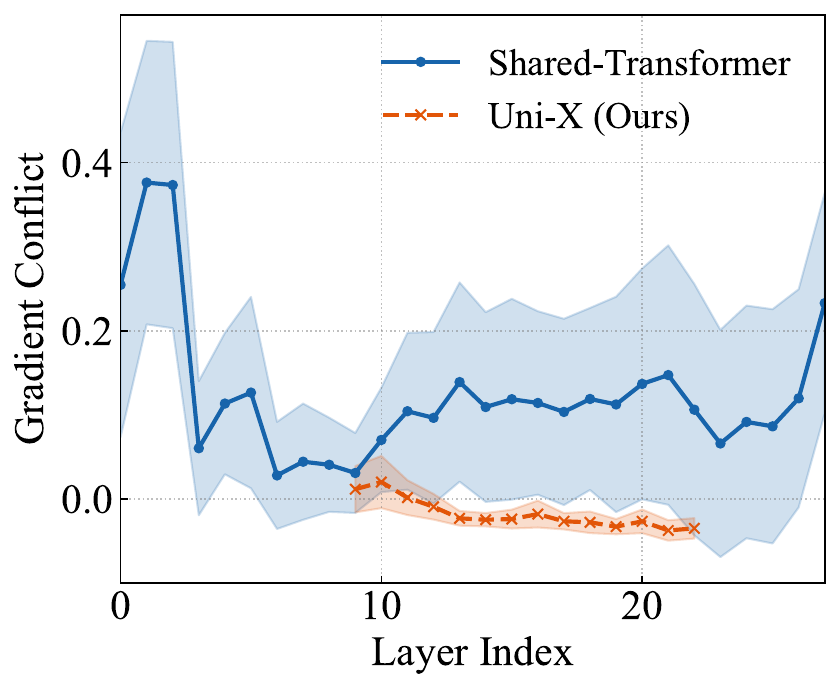}
      \vspace{-1.0em}
      \caption{Gradient conflict analysis of down-projection weights in the FFN of a modality-shared transformer. The shared transformer exhibits severe conflicts in shallow and deep layers, with only partial mitigation in intermediate layers. In contrast, Uni-X avoids conflicts at both extremes and further alleviates them in the middle layers.}
      \label{fig:grad_conflict}
      \vspace{-1.0em}
    \end{figure}
  \end{minipage}
\end{wrapfigure}

As illustrated in Figure~\ref{fig:grad_conflict}, these conflicts are most pronounced in shallow (input) and deep (output) layers due to differing text-image statistics.
In contrast, the middle layers, where representations become increasingly abstract and semantic~\citep{meng2022locating, geva_transformer_2021, sun2025transformerlayerspainters}, show reduced conflicts and stronger cross-modal alignment.
%%%
This suggests that an effective UMM should \textbf{respect modality-specific differences} rather than enforcing uniform parameter sharing across all layers.

Guided by this observation, we introduce \textbf{Uni-X}, a \emph{two-end-separated, middle-shared} architecture for unified multimodal modeling. 
%%%
In Uni-X, the shallow and deep layers are modality-specific, enabling specialized processing of distinct low-level distributions in text and vision, while the middle layers are shared to capture high-level semantic abstractions common to both.
%%%
This \textbf{X-shaped architecture} not only mitigates the severe gradient conflicts at the two ends but also further alleviates residual conflicts in the shared middle layers by leveraging natural semantic alignment between modalities (Figure~\ref{fig:grad_conflict}).

To demonstrate the effectiveness of Uni-X, we conduct extensive experiments under controlled training budgets and scaling regimes.
%%%
Results show that Uni-X improves training efficiency and achieves stronger performance under identical conditions. 
Moreover, with larger data and model scales, our 3B-parameter Uni-X matches or surpasses the performance of existing 7B AR-based UMMs across both understanding and generation benchmarks, demonstrating its scalability and competitiveness.
%%%
Our contributions are threefold:
\begin{itemize}[leftmargin=10pt] % nolistsep,
  \item \textbf{Empirical Analysis:} 
  We identify and quantify gradient conflicts between text and vision modalities in the shallow and deep layers of shared autoregressive transformers, attributing them to fundamental differences in their low-level statistical properties.
  
  \item \textbf{Model Design:} 
  We propose Uni-X, a novel two-end-separated, middle-shared architecture that aligns model structure with modality characteristics by using modality-specific layers for low-level processing and a shared core for high-level semantic fusion.
  
  \item \textbf{Comprehensive Validation:} Extensive experiments demonstrate that Uni-X improves training efficiency and scales effectively, enabling a 3B model to achieve performance competitive with much larger 7B models across diverse multimodal benchmarks.
\end{itemize}
\section{Related Work}
\label{sect:related}

\paragraph{Visual Language Models (VLMs).} 
The remarkable progress of LLMs~\citep{llama, yang2024qwen2_5, gpt3} has motivated researchers to extend them with visual cognition, giving rise to VLMs~\citep{llava, achiam2023gpt4}. 
Most VLMs leverage pre-trained visual encoders such as CLIP \citep{CLIP} or SigLIP2
\citep{tschannen2025siglip2multilingualvisionlanguage} to extract semantic features from images, which are projected into the LLM's semantic space via multimodal adapters~\citep{llava, liu2024llavanext, beyer2024paligemmaversatile3bvlm, team2024gemma, li2025ariaopenmultimodalnative}.
%%%
This design enables strong multimodal understanding and reasoning but remains \textbf{asymmetric}: VLMs treat images only as inputs and cannot generate them, limiting synergy between perception and synthesis.

\paragraph{Unified Multimodal Models (UMMs).} 
To enable such synergy, recent efforts have shifted toward UMMs, which aim to support both understanding and generation within a single framework~\citep{team_chameleon_2025, jin_unified_2024, wu2024vila}. 
A natural extension is to adopt the autoregressive (AR) paradigm of LLMs by treating visual tokens as a ``foreign language'' via vector quantization~\citep{wu_liquid_2025, wang2024emu3}.
%%%
However, the distinct statistical properties of text and images often lead to \textbf{modality conflicts}, degrading performance in shared transformers~\citep{team_chameleon_2025}.

To mitigate this, several approaches increase architectural complexity: 
Mixture-of-Transformers (MoT) designs~\citep{liao_mogao_2025, deng_emerging_2025, shi_lmfusion_2025} separate understanding and generation with distinct branches; 
Hybrid AR–diffusion frameworks~\citep{zhao2024monoformer, wu_omnigen2_2025, dreamllm, ge2025seedxmultimodalmodelsunified} combine next-token prediction with diffusion-based image synthesis; 
and branching strategies such as UniFork~\citep{li_unifork_2025} add task-specific deep heads. 
%%%
While effective on benchmarks, these methods sacrifice parameter sharing, complicate training, and weaken cross-modal benefits--the very goals UMMs were meant to unify.

\paragraph{Comparison with Uni-X.}
Uni-X builds on these insights but takes a different path. 
Instead of adding modules, Uni-X retains the simplicity of pure AR UMMs while mitigating modality conflict through a \textbf{two-end-separated, middle-shared} architecture: shallow and deep layers use modality-specific parameters to process low-level statistical differences, while intermediate layers are shared to exploit high-level semantic alignment.
%%%
This X-shaped design avoids the rigidity of MoT or UniFork and the complexity of AR-diffusion hybrids, offering a lightweight yet effective solution. We further discuss other related works in Appendix~\ref{appendix:other_related_work}.
% Empirically, Uni-X achieves comparable or superior performance to larger 7B AR UMMs across text, vision understanding, and generation tasks, while maintaining parameter efficiency and training simplicity. 
\section{Uni-X}
\label{sect:method}

\begin{figure}[t]
  \centering
  \includegraphics[width=0.99\textwidth]{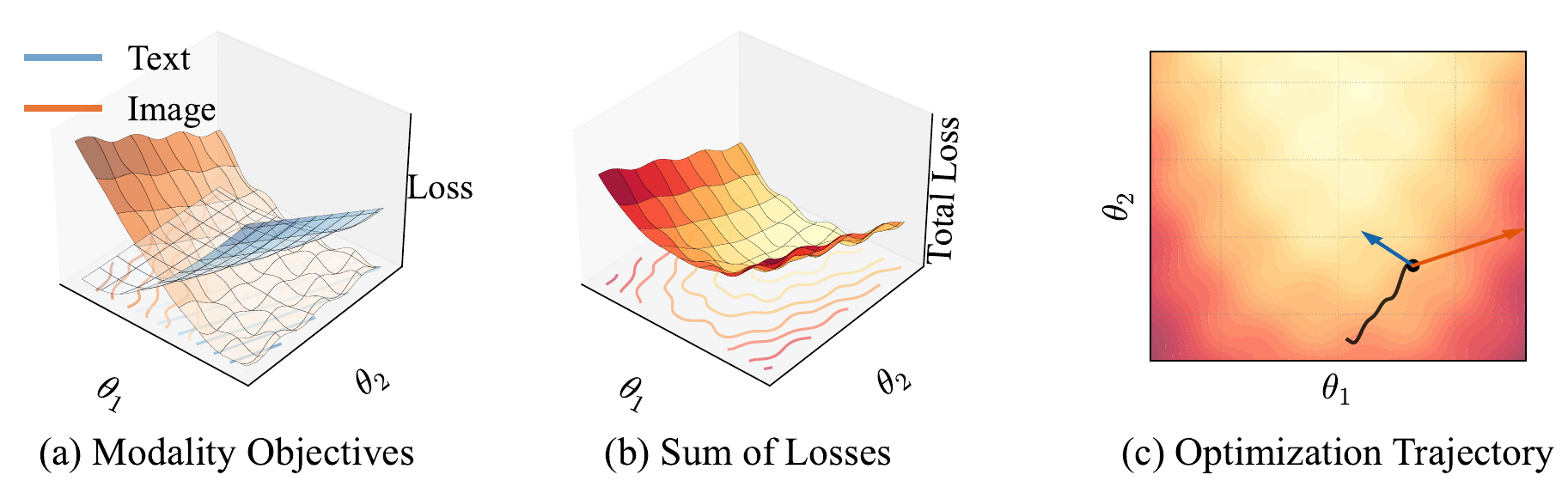}
  \vspace{-0.5em}
  \caption{Illustration of gradient conflict. (a) The loss landscapes of different modalities exhibit distinct geometries, creating potential conflicts in optimization direction. (b) The optimum of the sum of losses is different from the optimum of any single modality's loss. (c) In the presence of gradient conflict, the optimization trajectory becomes oscillating and suffers from slow convergence.}
  \label{fig:grad_conflict_illustration}
  \vspace{-1.0em}
\end{figure}

\subsection{Observations}
\label{sect:method:obser}

Before introducing the Uni-X architecture, we first analyze the gradient conflicts that arise when training modality-shared autoregressive transformers on multimodal data. We also provide an information-theoretic perspective to explain why such conflicts emerge.

\paragraph{Definition of Gradient Conflict.} 
As illustrated in Figure~\ref{fig:grad_conflict_illustration}, gradient conflict occurs when different optimization objectives induce gradients pointing in divergent directions, making joint optimization unstable and inefficient.
%%%
To quantify gradient conflict in multimodal training, we use an early checkpoint of a fully shared transformer. From this model checkpoint, we compute the \emph{average gradients} for specific parameter groups (e.g., FFN down-projection weights) using over 60 mini-batches and totaling 2M tokens, to ensure obtaining stable gradients.

Specifically, we first compute the average text gradient, $\bm{g}_{\texttt{text}}$. This is obtained by exclusively performing forward and backward passes on $D_{\texttt{text}}$, a text-only subset filtered from the pre-training data. 
%%%
Next, we compute the average image-text gradient, $\bm{g}_{\texttt{img}}$, using an analogous subset $D_{\texttt{img}}$ containing image-text pairs. 
%%%
The raw inter-modal similarity is then measured as the cosine similarity between these two average gradients:
\begin{equation}
    S_{\texttt{inter}} = \cos(\bm{g}_{\texttt{text}}, \bm{g}_{\texttt{img}}).
\end{equation}

However, since transformer layers have distinct roles at depth~\citep{sun2025transformerlayerspainters, geva2021transformer}, the resulting raw similarity $S_{\texttt{inter}}$ is biased and cannot be directly compared.
%%%
To correct for this, we estimate a baseline similarity $S_{\texttt{base}}$ that reflects the model's intrinsic gradient consistency on a unified data distribution. 
We randomly shuffle the full multimodal dataset $D_{\texttt{all}}$ and split it into two disjoint halves, $D_\texttt{any}^1$ and $D_\texttt{any}^2$. 
%%%
Their respective average gradients, $\bm{g}_{\texttt{any}}^1$ and $\bm{g}_{\texttt{any}}^2$, yield:
\begin{equation}
  S_{\texttt{base}} = \cos(\bm{g}_{\texttt{any}}^1, \bm{g}_{\texttt{any}}^2).
\end{equation}

This value represents the expected gradient similarity when gradients originate from the same underlying distribution.
Thus, we define the \textbf{gradient conflict} $c_g$ as the deviation from this baseline:
\begin{equation}
\label{eqn:gradient_conflict}
  c_g = -(S_{\texttt{inter}} - S_{\texttt{base}}).
\end{equation}

A high $S_{\texttt{base}}$ indicates the model's gradients are stable, whereas a much lower $S_{\texttt{inter}}$ suggests that the text-only and image-text data push the shared model parameters in conflicting directions, resulting in a large positive $c_g$. 
This provides a principled, layer-wise measure of inter-modal \emph{disagreement} and reveals where and why conflicts are most severe.

\paragraph{Empirical Findings.}
Figure~\ref{fig:grad_conflict} shows gradient conflict profiles ($c_g$) across depth. In modality-shared transformers, conflicts are most pronounced in shallow layers (near input) and deep layers (near output), while intermediate layers exhibit weaker conflicts. Experiments further reveal that Uni-X avoids conflicts at both extremes and reduces residual conflicts in the middle, validating its structural design. Additional analyses of other modules and the relationship between gradient conflict and data, as well as its impact on model performance, are provided in Appendix~\ref{appendix:more_grad_conflict}.

\paragraph{Why Do Conflicts Arise? Vision as a ``Foreign Language''} 
To explain these observations, we examine whether vision behaves like a ``foreign language'' when tokenized.
Using the VQ tokenizer~\citep{team_chameleon_2025}, images are represented as discrete token sequences, formally similar to text. 
Then, we define conditional entropy based on $n$-gram. 
When $n = 1$, the calculation reduces to ordinary information entropy. For $n > 1$, the conditional entropy is computed as follows:  
\begin{equation}
  H_{n} = -\sum p(w_{n} \mid w_{1}, w_{2}, \cdots, w_{n-1}) \log p(w_{n} \mid w_{1}, w_{2}, \cdots, w_{n-1}).
\end{equation}

\begin{wrapfigure}{r}{0.50\textwidth}
  \vspace{-2.5em}
  \begin{minipage}{\linewidth}
    \begin{figure}[H]
      \centering
      \includegraphics[width=0.95\columnwidth]{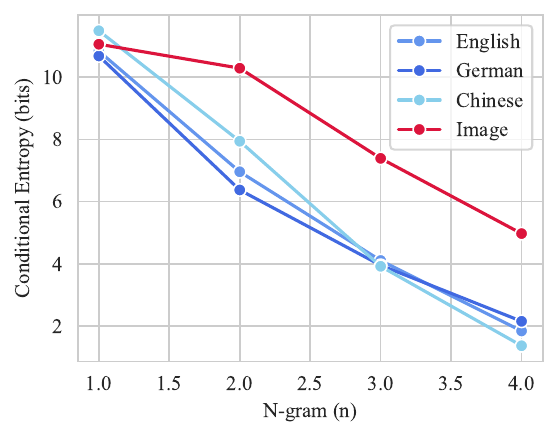}
      \vspace{-1.25em}
      \caption{Conditional entropy of images and natural languages. Image token sequences encoded by the VQ tokenizer exhibit substantially higher entropy, indicating greater difficulty in prediction.}
      \label{fig:cond_entropy}
      \vspace{-1.25em}
    \end{figure}
  \end{minipage}
\end{wrapfigure}
Results (Figure~\ref{fig:cond_entropy}) show that image tokens exhibit far higher entropy than natural languages such as English, German, or Chinese. 
While languages differ in grammar and lexicon, their token statistics remain closer to each other than to images. 
This means visual sequences are inherently harder to predict, requiring modeling of long-range, spatially entangled dependencies.

% However, as shown in Figure~\ref{fig:cond_entropy}, our conditional entropy analysis reveals significant differences between images and text. It is widely believed that the differences between Chinese and English are substantial because they belong to entirely different language families. However, after tokenization, the differences between languages are relatively small compared to the differences observed after the tokenization of images. Notably, due to computational constraints, we are unable to calculate conditional entropy for $n > 3$ in images, as this leads to an exponential growth of $n$-grams, approaching $3000^n$. 

As information theory suggests, sequences with higher (conditional) entropy are inherently harder to predict, requiring models to learn longer-range dependencies and more complex patterns. 
Thus, when a shared transformer jointly processes low-entropy, grammatical text with high-entropy, spatially complex vision, shallow and deep layers are forced to reconcile conflicting low-level distributions, producing strong gradient conflicts.
%%%
In contrast, intermediate layers, where representations become more abstract and semantic, naturally align across modalities, explaining the reduced conflicts observed in practice.

\begin{figure}[t]
  \centering
  \includegraphics[width=0.99\textwidth]{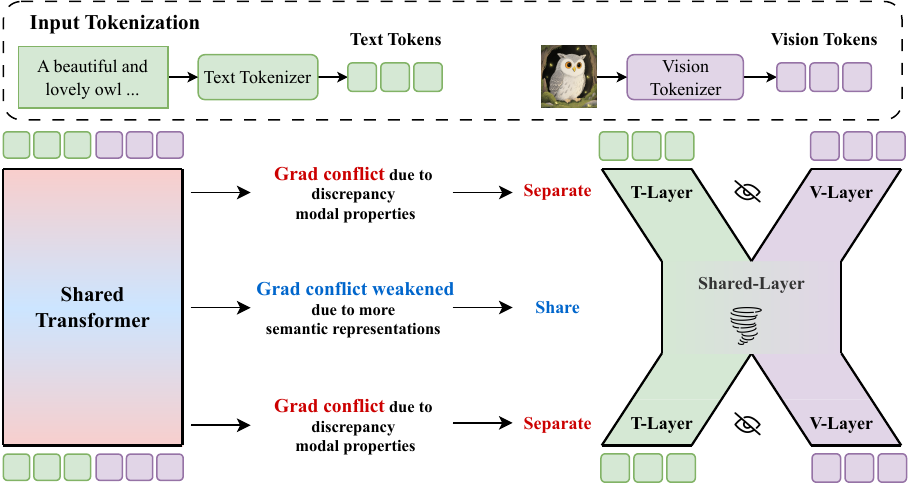}
  \vspace{-0.5em}
  \caption{Illustration of the proposed \textbf{Uni-X} architecture compared with a standard modality-shared transformer. 
  The baseline shared transformer (\textbf{left}) encounters gradient conflicts in shallow and deep layers due to the mismatched statistical properties of vision and text tokens.
  In contrast, Uni-X (\textbf{right}) adopts a two-end-separated, middle-shared design: modality-specific layers at both ends handle low-level feature processing, while a shared central block performs high-level semantic fusion.
  This structure aligns the architecture with the inherent characteristics of each modality and effectively mitigates gradient conflicts.}
  \label{fig:model_arch}
  \vspace{-0.5em}
\end{figure}

\subsection{Model Architecture}
\label{sect:method:arch}

Motivated by these findings, we propose \textbf{Uni-X}, an architecture designed to explicitly align model structure with modality characteristics.

\paragraph{Core Principle.}
As illustrated in Figure~\ref{fig:model_arch}, Uni-X follows a two-end-separated, middle-shared design. 
%%%
The shallow and deep layers are duplicated into parallel modality-specific branches, ensuring independent handling of text and vision during early feature extraction and final token projection. 
The intermediate layers remain shared, enabling high-level semantic fusion across modalities. 
%%%
This X-shaped separation-and-sharing balances modality specialization with semantic alignment.

\paragraph{Input Tokenization.}
For visual inputs, we employ the VQGAN tokenizer~\citep{esser2021taming} from Chameleon~\citep{team_chameleon_2025} to encode $512 \times 512$ images into a $32 \times 32$ grid of visual tokens from an 8,192-entry codebook. 
To accommodate these new tokens, we expand the vocabulary and corresponding embedding matrix of the base LLM. 
%%%
Textual inputs are processed via the standard BPE tokenizer. 
%%%
The final token sequence is structured as \texttt{<BOI>[Image]<EOI>[Text]<BOS>} for image understanding tasks and \texttt{[Text]<BOI>[Image]<EOI><BOS>} for image generation tasks. 
%%%
This unified tokenization enables AR training across modalities. 
While Uni-X can handle interleaved multimodal sequences, this study focuses on non-interleaved inputs.

\paragraph{Forward Propagation.}
Given a pre-trained LLM with $L$ layers, denoted as $\{\texttt{Layer}^i_\texttt{t}\}_{i=0}^{L-1}$, we partition them into three sections: 
The initial $N$ layers and the final $M$ layers constitute the ``separated layers,'' while the intermediate layers form the ``shared layers.'' 
Within the separated blocks, we introduce a new set of vision-specific layers, $\{\texttt{Layer}^i_\texttt{v}\}$, which operate in parallel with the original text layers, $\{\texttt{Layer}^i_\texttt{t}\}$.

To manage the data flow, we introduce a binary mask $\bm{M}_\texttt{v} \in \{0, 1\}^n$ to identify the positions of visual tokens. At any given layer $l$, the complete hidden states $\bm{H}^l$ can be partitioned into text-specific states $\bm{H}^l_\texttt{t} = \bm{H}^l[\sim \bm{M}_\texttt{v}]$ and vision-specific states $\bm{H}^l_\texttt{v} = \bm{H}^l[\bm{M}_\texttt{v}]$.

The forward propagation in Uni-X is defined as follows:
\begin{equation}
  \bm{H}^{l+1}_x = \begin{cases} 
    \texttt{Layer}^l_x(\bm{H}_x^l) & \text{if } l < N \text{ or } l \geq L - M, \\ 
    \left[ \texttt{Layer}^l_\texttt{t}(\bm{H}^l) \right]_x & \text{otherwise,}
  \end{cases}
\end{equation}
where $x \in \{\texttt{t}, \texttt{v}\}$ denotes the modality (text or vision). In the ``otherwise'' case, $[\cdot]_x$ indicates selecting the subset of the output hidden states corresponding to modality $x$.

Importantly, unlike other architectures~\citep{li_unifork_2025, deng_emerging_2025, shi_lmfusion_2025}, the vision and text modalities remain strictly isolated within the separated blocks, with no cross-modal interaction. This forces the model to learn robust unimodal representations before they are fused in the shared block and after they are separated for modality-specific output generation. 
% We hypothesize this prevents premature and potentially detrimental cross-modal entanglement. \todo{An experiment is needed to validate this hypothesis.}

\paragraph{Training Objective.}
Following the standard paradigm for AR models, Uni-X is trained to predict the next token in a sequence containing both text and visual tokens. 
%%%
The training objective is to minimize the cross-entropy loss over the vocabulary for each token. The loss function $\mathcal{L}$ for a given sequence $\mathcal{S} = (s_1, s_2, \cdots, s_T)$ is defined as:
\begin{equation}
  \mathcal{L} = -\sum_{i=1}^{T} \log P(s_i \mid s_{<i}).
\end{equation}
This simple yet effective objective enables the model to learn both understanding and generation capabilities across modalities within a single, unified framework.

\paragraph{Design Rationale.}
Unlike prior architectures that rely on auxiliary semantic encoders~\citep{wu_janus_2024, chen_janus-pro_2025}, hybrid AR–diffusion pipelines~\citep{wu_omnigen2_2025, zhao2024monoformer}, or task-specific branching structures such as MoT~\citep{liao_mogao_2025} and UniFork~\citep{li_unifork_2025}, Uni-X maintains the simplicity of a pure autoregressive framework. 
%%%
Its two-end-separated, middle-shared structure is motivated directly by empirical evidence of gradient conflicts, aligning the model design with the statistical characteristics of each modality. 
By isolating low-level modality-specific processing while preserving a shared semantic core, Uni-X avoids the complexity and overhead of multi-expert or dual-paradigm systems, yet achieves competitive or superior performance. 
This balance of architectural simplicity, empirical grounding, and scalability makes Uni-X a practical foundation for unified multimodal modeling.

\section{Experiments}
\label{sect:expt}

We evaluate Uni-X from two complementary perspectives:
(1) \textbf{Efficiency under identical training conditions}, where Uni-X and baseline architectures are trained on the same data and resources, enabling fair comparisons of efficiency and performance.
(2) \textbf{Scaling within resource constraints}, where we maximize dataset size and training duration to examine Uni-X’s scalability and competitiveness against larger state-of-the-art models.

\subsection{Experimental Setup}
\label{sect:expt:setup}

\paragraph{Pre-training Datasets.}
Our pre-training stage was designed to build a strong foundation in both language and vision. 
%%%
To preserve general text generation capabilities, we utilized a diverse set of text corpora: the high-quality Chinese dataset CCI3-H~\citep{wang2024cci30hqlargescalechinesedataset}, English datasets DCLM~\citep{li2025datacomplmsearchgenerationtraining} and Fineweb-Edu~\citep{penedo2024fineweb}, and the StarcoderData~\citep{li2023starcodersourceyou} corpus, as integrating code is known to boost general model performance~\citep{ma2024at}. 
%%%
For multimodal pre-training, we used public benchmarks like ImageNet~\citep{russakovsky2015imagenetlargescalevisual} and JourneyDB~\citep{pan2023journeydb}, complemented by a substantial internally collected dataset of 40 million images, which were captioned using the powerful Intern-VL model~\cite{internvl1.5}. 
%%%
Following the methodology of Liquid~\citep{wu_liquid_2025}, we diversified our training data by randomly reversing 20\% of the text-to-image pairs to serve as image-captioning tasks. The final pre-training data consists of 72B text tokens and 65B vision tokens.

\paragraph{Supervised Fine-Tuning (SFT).}
We further refined the model with 3B SFT tokens. 
For vision understanding, we employed MiniGemini~\citep{li_mini-gemini_2024} and FineVision~\citep{huggingface2025finevision}. 
To improve text understanding and general instruction-following, we utilized OpenOrca~\citep{openorca}. 
Additionally, to refine the quality of image generation, we leveraged Blip3o-60k~\citep{chen_blip3-o_2025} and ShareGPT4o~\citep{chen2023sharegpt4v}.

\paragraph{Benchmarks.}
Evaluation covered text-only, image generation, and multimodal understanding tasks. 
%%%
For text-only tasks, we employed ARC-Easy/Challenge (\textbf{ARC-E}/\textbf{ARC-C})~\citep{arc_clark2018thinksolvedquestionanswering}, WinoGrande (\textbf{WinoG})~\citep{sakaguchi2020winogrande}, \textbf{BoolQ}~\citep{clark2019boolq}, and \textbf{MMLU}~\citep{mmlu_hendrycks2021measuringmassivemultitasklanguage}.
%%%
For image generation, we used \textbf{GenEval}~\citep{ghosh2023genevalobjectfocusedframeworkevaluating} and DPG-Bench (\textbf{DPG})~\citep{dpg_bench}. 
For the GenEval benchmark, we followed Bagel~\citep{deng_emerging_2025} and employed an LLM to rewrite shorter prompts into more detailed ones to better assess instruction following.
%%%
For multimodal understanding, we used SEEDBench (\textbf{SEED})~\citep{seedbench}, \textbf{MME}~\citep{mme}, \textbf{POPE}~\citep{POPE}, and MMBench (\textbf{MMB})~\citep{liu2024mmbenchmultimodalmodelallaround}.  

\paragraph{Implementation Details.}
We conducted ablation studies on Qwen2.5-1.5B~\citep{yang2024qwen2_5} and scaled to Qwen2.5-3B. 
%%%
We used the VQGAN tokenizer~\citep{esser2021taming} from Chameleon~\citep{team_chameleon_2025} to encode $512 \times 512$ images into $32\times 32$ discrete tokens. 
Our codebase is built upon the Liquid~\citep{wu_liquid_2025} and HuggingFace Transformers~\citep{wolf2019huggingface} libraries. 
Training was accelerated using Flash Attention 2~\citep{dao2022flashattention} and DeepSpeed ZeRO2~\citep{aminabadi2022deepspeed}. 
When generating images, we uniformly set the classifier-free guidance (CFG) to 4.0.

\begin{table}[tb]
\centering
\small
\renewcommand{\arraystretch}{1.2}
\caption{The text performance of Uni-X compared to other models. }
\label{tab:main_text}
\vspace{0.5em}
\resizebox{\columnwidth}{!}{%
% \adjustbox{width=0.9\textwidth,totalheight=\textheight,keepaspectratio}{
\begin{tabular}{ll|rrrrrr}
\toprule
\rowcolor[HTML]{DDEBFF} \textbf{Model} &  \textbf{\# Params.} &\textbf{ARC-E} & \textbf{ARC-C} & \textbf{WinoG} & \textbf{BoolQ} & \textbf{MMLU} & \textbf{Avg. $\uparrow$} \\
\midrule
\textbf{Janus-Pro}~\citep{chen_janus-pro_2025}&  7B&70.4& 40.9& 66.1& 80.2& 49.3& 61.4\\
\textbf{VILA-U}~\citep{wu2024vila}&  7B&51.6& 34.0& 57.3& 70.6& 25.5& 47.8\\
\textbf{Chameleon}~\citep{team_chameleon_2025}&  7B&76.1& 46.5& 70.4& 81.4& 52.1& 65.3\\
% \textbf{Chameleon-34B}&  34B&84.1& 59.7& 78.5& 86.0& 65.8& 74.8\\
\textbf{Liquid}~\citep{wu_liquid_2025}&  7B&75.6& 49.0& 72.7& 81.0& 56.0& 66.9\\
\midrule
\rowcolor[HTML]{FFF2CC} \textbf{Uni-X} &  3B / 4.5B&79.0& 47.9& 68.9& 82.2& 57.6& 67.1\\
% \rowcolor[HTML]{FFF2CC} \textbf{Uni-X}$^\heartsuit$ &  3B / 4.5B&79.0& 47.9& 68.9& 82.2& 57.6& 67.1\\
\bottomrule
\end{tabular}}
\end{table}
\setlength{\textfloatsep}{1.5em}

\begin{table}[t]
\centering
\small
\vspace{-1.0em}
\renewcommand{\arraystretch}{1.15}
\caption{The image generation and multimodal understanding performance of Uni-X compared to other models. In the \textbf{\# Params} column $x/y$, $x$ and $y$ represent the number of active parameters and the total parameters, respectively. $^{\dagger}$ represents the model variant that performs semantic alignment. $^\ddagger$ represents the rewriting of the prompt during evaluation. $^\heartsuit$ indicates that it has been trained on more image-text data.}
\label{tab:main_img}
\vspace{0.5em}
\resizebox{\columnwidth}{!}{%
\begin{tabular}{lll|lr|rrrr}
  \toprule
  \rowcolor[HTML]{DDEBFF} \textbf{Model} & \textbf{\# Tokens} & \textbf{\# Params.} & \textbf{GenEval} & \textbf{DPG} & \textbf{MME}& \textbf{POPE} & \textbf{MMB} & \textbf{SEED} \\
  \midrule
  
  %--- Category for Autoregressive meets Diffusion Models ---
  \multicolumn{9}{l}{\textit{\textcolor{gray}{Autoregressive meets Diffusion}}} \\
  \midrule
  \textbf{Bagel}~\citep{deng_emerging_2025}   & 5.1T & 7B / 14B     & 88$^\ddagger$& 85.0  & -         & -     & 85.0  & -    \\
  \textbf{Bilp3o}~\citep{chen_blip3-o_2025}  & - & 4B / 9B & 81& 79.3  & 1,527.7    & -     & 78.6  & 73.8 \\
  \textbf{X-Omni}~\citep{geng_x-omni_2025}  & \textasciitilde 1T  & 10B / 20B  & 83$^\ddagger$& 87.6  & -  & 89.3  & 74.8  & 74.1 \\
  \textbf{Show-o}~\citep{xie2024show} & \textasciitilde 500B  & 1.3B  & 68 & - & 1,097.2 & 80.0 & - & - \\
  \textbf{Show-o}$^{\dagger}$~\citep{xie2024show} & \textasciitilde 500B & 1.3B & 69& - & 1,232.9 & 84.5 & - & -\\
  \midrule
  
  %--- Category for Autoregressive w/ Semantic Encoder ---
  \multicolumn{9}{l}{\textit{\textcolor{gray}{Autoregressive w/ Semantic Encoder}}} \\
  \midrule
  \textbf{NextStep1}~\citep{team_nextstep-1_2025}      & \textasciitilde 1T      & 14B & 73$^\ddagger$& 85.2  & -  & -  & -     & -    \\
  \textbf{Janus-Pro}~\citep{chen_janus-pro_2025} & \textasciitilde 300B    & 7B    & 80& 84.1  & -   & 87.4  & 79.2  & 72.1 \\
  \textbf{VILA-U}~\citep{wu2024vila} &  -       & 7B           & -     & -     & 1,336.2    & 83.9  & -     & 56.3 \\
  \textbf{Liquid}$^{\dagger}$~\citep{wu_liquid_2025} &  -       & 8B     & -     & -  & 1,448.0 & 83.2 & -  & -\\
  \midrule

%--- Category for Autoregressive w/o Semantic Encoder ---
\multicolumn{9}{l}{\textit{\textcolor{gray}{Autoregressive w/o Semantic Encoder}}} \\
\midrule
\textbf{Chameleon}~\citep{team_chameleon_2025}   & 9.2T                    & 34B          & 39& -     & 604.5     & -     & 32.7  & -    \\
% \textbf{EVE} (HD)& \textasciitilde 100B                    & 7B          & -  & -     & 1305.7& 85.0& 52.3& 56.8\\
\textbf{LWM}~\citep{liu2024world_lwm}         & \textasciitilde 500B    & 7B           & 47& -     & -         & 75.2  & -     & -    \\
\textbf{EMU3}~\citep{wang2024emu3}        & -                       & 8B           & 66$^\ddagger$& 80.6  & 1,243.8    & 85.2  & 58.5  & 68.2 \\
\textbf{Liquid}~\citep{wu_liquid_2025}      & \textasciitilde 90B     & 7B           & 68$^\ddagger$& 79.8  & 1,107.2    & 81.1  & -     & -    \\
\midrule
\rowcolor[HTML]{FFF2CC} \textbf{Uni-X} & 140B & 3B / 4.5B & 82$^\ddagger$& 79.8 & 1,158.3 & 83.6 & 59.3 & 60.2\\
\rowcolor[HTML]{FFF2CC} \textbf{Uni-X}$^\heartsuit$ & 240B & 3B / 4.5B & 83$^\ddagger$& 80.3 & 1,228.2 & 84.6 & 62.7 & 59.8\\
\bottomrule
\end{tabular}}
\end{table}

\begin{table}[t]
\centering
\renewcommand{\arraystretch}{1.3}
\caption{Image edit results on ImgEdit-Bench. $^\heartsuit$ denotes models trained on extra image-text data.}
\label{tab:img_edit}
% 将表格缩放到完整的文本宽度
\vspace{0.5em}
\resizebox{\columnwidth}{!}{%
% \adjustbox{width=0.99\textwidth,totalheight=\textheight,keepaspectratio}{
\begin{tabular}{ll | rrr rrr rrr r}
    \toprule
    \rowcolor[HTML]{DDEBFF} \textbf{Model} & \textbf{\# Params.} & \textbf{Add} & \textbf{Adjust} & \textbf{Extract} & \textbf{Replace} & \textbf{Remove} & \textbf{Background} & \textbf{Style} & \textbf{Hybrid} & \textbf{Action} & \textbf{Overall\(\uparrow\)} \\
    \midrule
    \textbf{GPT-4o}     & -         & 4.61 &   4.33 &    2.90 &    4.35 &   3.66 &       4.57 &  4.93 &   3.96 &   4.89 &          4.20 \\
    \textbf{ICEdit}      & 12B       & 3.58 &   3.39 &    1.73 &    3.15 &   2.93 &       3.08 &  3.84 &   2.04 &   3.68 &              3.05 \\
    \textbf{AnyEdit}     & 4B        & 3.18 &   2.95 &    1.88 &    2.47 &   2.23 &       2.24 &  2.85 &   1.56 &   2.65 &              2.45 \\
    \textbf{UltraEdit}   & 4B        & 3.44 &   2.81 &    2.13 &    2.96 &   1.45 &       2.83 &  3.76 &   1.91 &   2.98 &              2.70 \\
    \textbf{Step1X-Edit} & 12B       & 3.88 &   3.14 &    1.76 &    3.40 &   2.41 &       3.16 &  4.63 &   2.64 &   2.52 &              3.06 \\
    \textbf{Bagel}       & 7B / 14B  & 3.56 &   3.31 &    1.70 &    3.30 &   2.62 &       3.24 &  4.49 &   2.38 &   4.17 &              3.20 \\
    \midrule
    \rowcolor[HTML]{FFF2CC} \textbf{Uni-X}$^\heartsuit$   & 3B / 4.5B & 3.57 &   3.18 &    2.06 &    3.94 &   3.82 &       3.38 &  4.21 &   3.16 &   3.63 &       3.44 \\
    \bottomrule
\end{tabular}}
\end{table}

\subsection{Results and Analysis}
\label{sect:expt:analysis}

\paragraph{Scaling Experiment.}
In this experiment, we aim to demonstrate that Uni-X can scale effectively and is not limited to small-scale training data. 
We expand the dataset size to $140$B total tokens for Uni-X, using Qwen2.5-3B as the base model, and extend the training duration to achieve improved performance. The evaluation is conducted against SOTA models, some of which have been trained on trillions of tokens.
%%%
As presented in Table~\ref{tab:main_text}, Uni-X robustly maintains the strong language capabilities of its base model. 
With an average score of 67.1 across five text benchmarks, our 3B Uni-X model outperforms several larger 7B models. 
This demonstrates that our design successfully mitigates modality conflict without sacrificing performance on fundamental language understanding tasks.

For image generation, as detailed in Table~\ref{tab:main_img}, Uni-X achieves a strong score of 82 on GenEval, a result that surpasses many models with more parameters and underscores the effectiveness of our architecture in producing high-quality images. We also tested T2I-CompBench~\citep{t2i_compbench} and MSCOCO~\citep{mscoco}, as shown in Appendix~\ref{appendix:more_results_on_img_gen}. Uni-X similarly exhibited strong performance with fewer parameters.
%%%
Regarding vision understanding (Table~\ref{tab:main_img}), while Uni-X's scores are slightly lower than some state-of-the-art models, we observe a clear trend: 
models that incorporate an additional semantic image encoder, such as Janus-Pro \citep{chen_janus-pro_2025} and the semantically aligned variants of Liquid \citep{wu_liquid_2025} and Show-o \citep{xie2024show}, tend to achieve substantially higher performance on understanding benchmarks like MMBench and SEED.

In contrast, among models that do not rely on a separate semantic encoder, Uni-X's performance is commendable and holds its ground against strong competitors like EMU3 \citep{wang2024emu3}. 
This suggests that our architecture effectively harnesses the inherent capabilities of the autoregressive framework for vision understanding. 
%%%
We speculate that the relatively weaker understanding performance might be partially caused by the insufficient utilization of the VQ tokenizer's codebook. We analyzed the token sequences encoded from 1 million images and found that, although there are 8,192 tokens available in the tokenizer, only $\approx$3,127 are being utilized. 
Meanwhile, EMU3 uses $4096$ tokens to represent a $512\times512$ image, which provides more fine-grained information. However, this $4\times$ token count severely impacts its image generation speed, as shown in Appendix~\ref{appendix:infer_eff}.

For image editing, we conducted tests on ImgEdit~\citep{img_edit} as shown in Table~\ref{tab:img_edit}. Uni-X achieved better results than Bagel, even with less training data and fewer parameters than Bagel. This demonstrates that the high-level semantic unification of Uni-X enhances its image editing capabilities.

\begin{table}[t]
\centering
\small
\vspace{-1.0em}
\renewcommand{\arraystretch}{1.1}
\caption{Performance and training efficiency comparison of different model architectures under identical training conditions. Training efficiency is measured by the number of tokens processed per second per GPU. $^{\spadesuit}$ indicates that the baseline has been adapted for our experimental setting (see Appendix~\ref{appendix:baseline_impl} for specific details); $^{\diamondsuit}$ represents calculating the loss on the instruction part during the training of image-text data. }
\label{tab:ablation_arch}
\vspace{0.5em}
\begin{tabular}{ll|rrrrr}
\toprule
\rowcolor[HTML]{DDEBFF} \textbf{Model} &  \textbf{\#Params.} & \textbf{MMLU}  &\textbf{GenEval}& \textbf{MMB} & \textbf{Avg.} $\uparrow$ & \textbf{Efficiency} $\uparrow$\\
\midrule
\textbf{Shared Transformer}&  1.5B &  50.0 &33.6 & 30.3 & 38.0 & 16,380 \\
\textbf{MoT}$^\spadesuit$~\cite{deng_emerging_2025}&  1.5B / 3B& 48.0&26.0& 30.0& 34.6& 12,658\\
\textbf{HardMoE}&  1.5B / 2.3B& 50.3 &42.8& 30.7& 41.3 & 14,657 \\
\textbf{UniFork}$^{\spadesuit}$~\citep{li_unifork_2025}&  1.5B / 2.3B&  50.1 &12.4& 25.9 & 29.5 & 15,481 \\
\midrule
\rowcolor[HTML]{FFF2CC} \textbf{Uni-X} (9:5)$^{\diamondsuit}$ &  1.5B / 2.3B & 48.5 &34.8 & 29.8 & 37.7 & 15,642 \\
\rowcolor[HTML]{FFF2CC} \textbf{Uni-X} (9:5)&  1.5B / 2.3B& 50.1 &43.3& 31.5& 41.6 & 15,595 \\
\bottomrule
\end{tabular}
\end{table}

\paragraph{Identical Training Conditions.} 
To validate the effectiveness of Uni-X, we conducted ablation experiments on a smaller dataset and a slightly reduced base model Qwen2.5-1.5B, due to resource constraints. To ensure consistency in performance comparisons, we limited the dataset to $28$B tokens, of which $13.7$B are vision tokens. The experiments were conducted using a learning rate (LR) of $5 \times 10^{-5}$, warmup ratio 0.03, and constant LR scheduler, with batch size 17,560 tokens per GPU.

The selected baselines include: 
(1) \textbf{Shared Transformer}, which continues multimodal pre-training based on Qwen2.5-1.5B; 
(2) \textbf{Mixture-of-Transformers (MoT)}~\citep{deng_emerging_2025}, where prior work replicates an additional transformer to handle image generation tasks, while the original LLM backbone focuses on text-only and image understanding tasks. Under our experimental setup, vision tokens are allocated to the duplicated transformer;
(3) \textbf{Hard-Route MoE (HardMoE)}, which introduces a vision expert specifically for the vision modality, assigning vision tokens to this expert for computation guided by the vision mask; and 
(4) \textbf{UniFork}~\citep{li_unifork_2025}, which creates a task-specific deep branch for image generation. 
%%%
For all the baselines, we ignored the instruction during training to enhance cross-modal performance and ensure a fair comparison.

Results (Table~\ref{tab:ablation_arch}) show that Uni-X achieves the best overall performance under consistent training conditions. 
Specifically, our Uni-X (9:5) configuration attains an average score of 41.6, significantly outperforming the standard baselines. 
While HardMoE is competitive, achieving a score of 41.3, Uni-X still holds a slight advantage. Moreover, HardMoE and UniFork are orthogonal and can be combined. 
In terms of training efficiency, although the baseline shared transformer is the fastest due to having the fewest parameters, Uni-X achieves a high throughput, which is considerably more efficient than the less performant MoT architecture. These findings confirm that Uni-X's design offers a more effective trade-off between performance and computational efficiency.

It is worth noting that the architectures of MoT and UniFork have been adapted to fit our VQ+AR setup to avoid discrepancies in efficiency between paradigms such as diffusion and AR+diffusion. Specific details can be found in Appendix~\ref{appendix:baseline_impl}. A comparison of training efficiency across paradigms lies beyond the scope of this work and will be considered in future research.

\begin{figure}[t]
  \centering
  \includegraphics[width=0.99\textwidth]{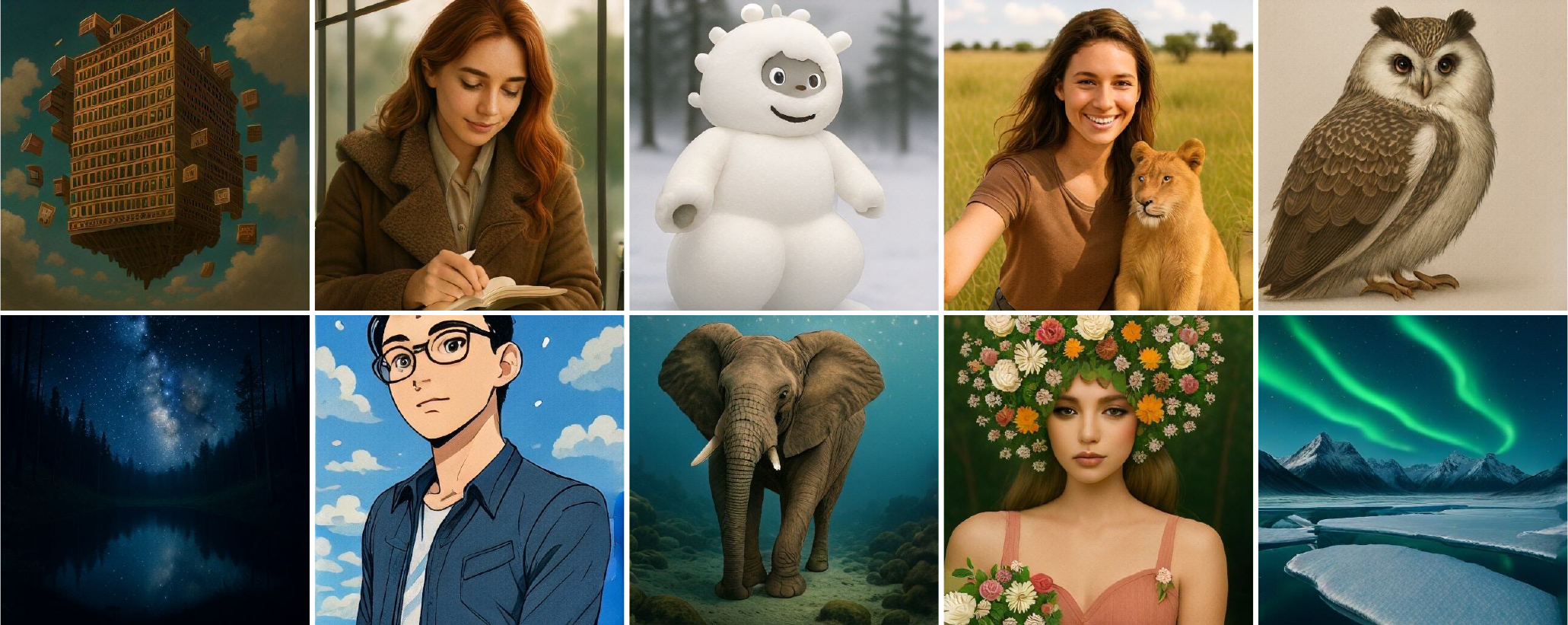}
  \vspace{-0.5em}
  \caption{Qualitative examples of Uni-X image generation. The results highlight its ability to produce diverse, high-quality visuals that follow prompts with both creativity and fine-grained detail.}
  \label{fig:gen_case}
  \vspace{-0.5em}
\end{figure}

\begin{wrapfigure}{r}{0.6\textwidth} % {位置}{宽度}
  \vspace{-1.0em}
  \centering
  % 建议将图片宽度改为 \linewidth 以自适应 wrapfigure 的宽度，或者保持原 absolute width
  \includegraphics[width=0.99\linewidth]{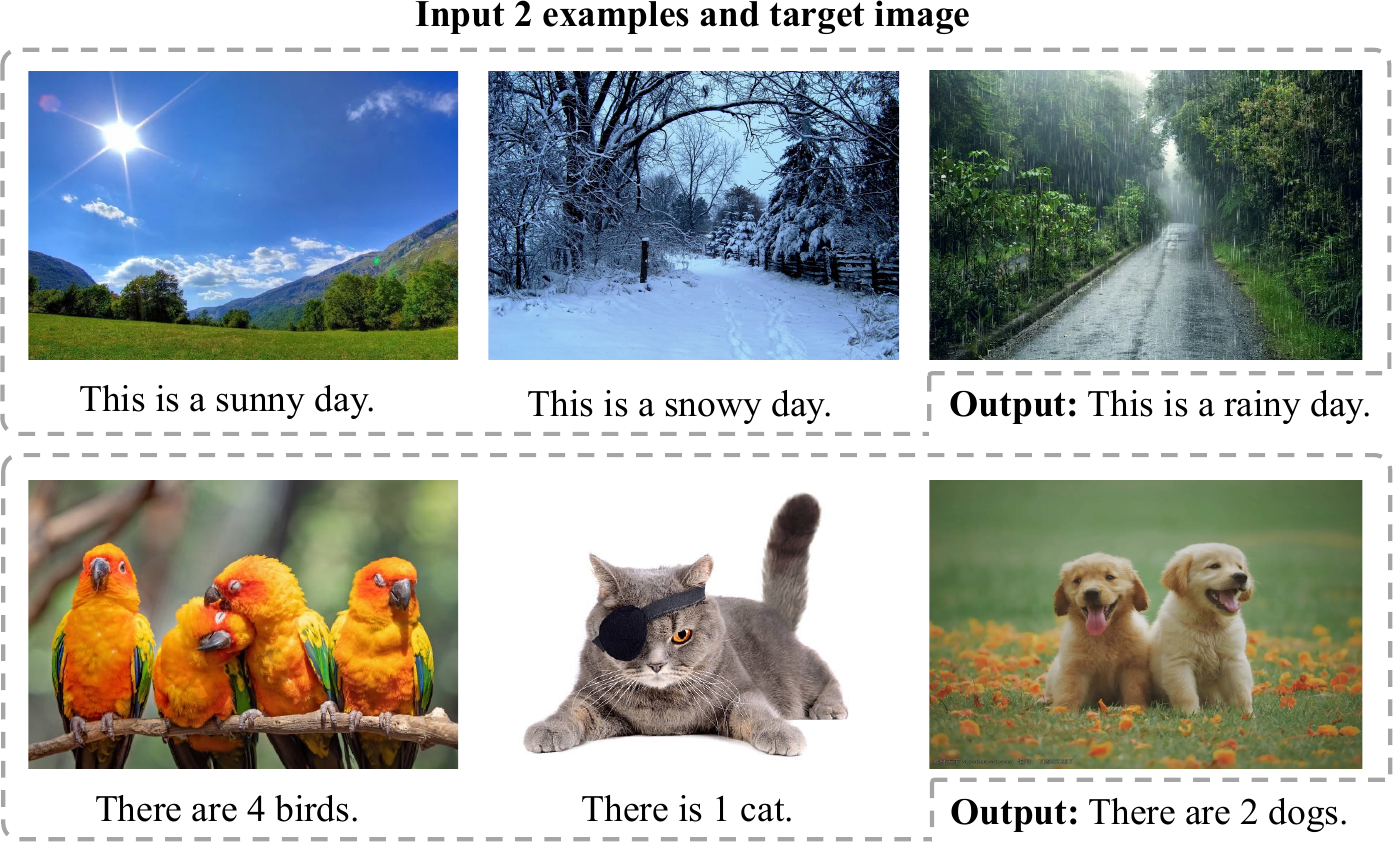}
  \vspace{-0.5em}
  \caption{Demonstration of Uni-X's in-context learning. The model follows few-shot examples to perform tasks such as image description (1st line) and object counting (2nd line).}
  \label{fig:in_context_learning}
  \vspace{-0.5em}
\end{wrapfigure}

\paragraph{Case Study.} 
In Figure ~\ref{fig:gen_case}, we present a curated selection of images generated by Uni-X to qualitatively assess its capabilities.
Despite the relatively limited number of tokens used during training, the model demonstrates a strong ability to produce clear, aesthetically pleasing images that exhibit robust instruction-following capabilities.
The examples showcase Uni-X's versatility in handling a wide range of creative and complex prompts.
For instance, the model can generate imaginative fantasy scenes, such as a gigantic library floating above the clouds, and surreal compositions, like a realistic elephant walking on the ocean floor.
Furthermore, Uni-X successfully adheres to specific artistic style requests, as seen in the detailed anime-style portrait, and renders fine details with high fidelity, exemplified by the intricate feather patterns of the owl.
These case studies verify that Uni-X can effectively translate complex textual descriptions into high-quality visual outputs.
The specific prompts used for these generations are provided in Appendix~\ref{appendix:prompts_gen_img}.

\paragraph{In-Context Learning.} 
Although Uni-X was not explicitly trained on interleaved multimodal data, we conducted an evaluation to assess its emergent in-context learning (ICL) capabilities. As illustrated in Figure~\ref{fig:in_context_learning}, the model was presented with few-shot examples, where several image-text pairs were provided as context before a final query image was presented without its corresponding description.

The results demonstrate that Uni-X can successfully interpret the contextual examples and apply the learned pattern to the target image. For instance, in the top row of Figure~\ref{fig:in_context_learning}, the model correctly identifies the weather in the target image as a ``rainy day,'' adhering to the simple descriptive format (``This is a... day.'') established by the preceding examples. Also, Uni-X exhibits the ability to perform more reasoning tasks, such as object counting. This suggests that the model is not merely mimicking sentence structure but is performing cross-modal reasoning at a semantic level.

\paragraph{Ignore Instruction in Training.} 
Ignoring the loss of the instruction part during training is a common technique in supervised fine-tuning. However, its role in pretraining is rarely emphasized. Following Liquid~\citep{wu_liquid_2025}, we applied the same ``ignore instruction'' strategy during pretraining. 

Specifically, no loss mask was applied for pure text data. For text-image pairs, in text-to-image tasks, the loss calculation excluded the text instruction tokens. Similarly, for image captioning tasks, the loss corresponding to the image tokens was masked. As demonstrated in our experimental results, Table~\ref{tab:ablation_arch}, this approach significantly enhanced the model's capability to generate images.

\begin{wraptable}{r}{0.6\textwidth}
\vspace{-2.0em} 
\centering
\caption{Performance comparison of different Uni-X configurations. Here, $x:y$ denotes the number of shallow separated layers $x$ and deep separated layers $y$, respectively. The total number of layers is $n=28$. The split points are $x$ and $n-y$, respectively.}
\label{tab:uni_x_configs}
\vspace{0.5em}
\adjustbox{width=0.9\linewidth}{
\begin{tabular}{lrrrr}
  \toprule
  \rowcolor[HTML]{DDEBFF} \textbf{Configuration} &  \textbf{MMLU} &\textbf{GenEval} & \textbf{MMB} & \textbf{Avg.~$\uparrow$} \\
  \midrule
  \textbf{Uni-X} (3:3)  &  48.7 &37.3 & 30.7 & 38.9 \\
  \textbf{Uni-X} (7:7)  &  49.6 &41.3 & 29.4 & 40.1 \\
  \textbf{Uni-X} (11:11)&  49.7 &37.5 & 32.1 & 39.8 \\
  \midrule
  \textbf{Uni-X} (3:11) &  50.0 &32.9 & 31.0 & 38.0 \\
  \textbf{Uni-X} (5:9)  &  50.1 &39.2 & 28.0 & 39.1 \\
  \textbf{Uni-X} (9:5)  &  50.1 &43.3 & 31.5 & 41.6 \\
  \textbf{Uni-X} (11:3) &  49.8 &25.1 & 31.9 & 35.6 \\
  \bottomrule
\end{tabular}}
\vspace{-0.5em}
\end{wraptable}

We believe there might be several reasons for this: 1) This mask forces the model to learn the relationship between the two modalities rather than relying on the prior distribution of images, thereby enhancing its instruction-following capability. 2) It serves as a form of loss regularization. For text-image pair data, the number of image tokens is fixed at 1024, while the average number of text tokens is around 120. By masking, we ensure that the gradient magnitude generated by the loss is only dependent on the reverse ratio we set.

\paragraph{Number of Separated Layers.}
We investigate how the number and distribution of separated layers affect performance (Table~\ref{tab:uni_x_configs}).
Varying the total number of separated layers produces an $n$-shaped trend: more separation improves modality-specific low-level processing, but too many layers reduce shared middle layers, weakening semantic fusion and cross-modal reasoning. 
The best overall performance is achieved with 14 separated layers.

We then examine shallow-deep ratios under this setting. A 9:5 split (slightly more shallow than deep layers) performs best, indicating that early processing of low-level features, where text and vision differ most, benefits more from modality-specific capacity than the final generation stage. These results provide strong empirical support for the Uni-X design. 
%%%
We also explored text layers and vision layers with different numbers of separate layers, and the results are shown in Appendix~\ref{appendix:tv_ratio}.

\section{Conclusions}
\label{sect:conclusion}

In this work, we identified gradient conflicts as a fundamental limitation of shared AR UMMs, particularly in the shallow and deep layers where vision and text exhibit highly divergent low-level statistics.
%%%
To address this challenge, we proposed Uni-X, a two-end-separated, middle-shared architecture that explicitly aligns model structure with modality characteristics. By isolating low-level processing into modality-specific branches while maintaining a shared semantic core for high-level fusion, Uni-X effectively mitigates inter-modal conflicts without adding architectural complexity.
%%%
Extensive experiments show that this X-shaped design allows a 3B-parameter Uni-X model to deliver performance competitive with much larger 7B UMMs across diverse multimodal benchmarks. 
These findings establish Uni-X as both a scalable and parameter-efficient foundation, paving the way for future research in unified multimodal modeling.

\section*{Acknowledgments}

We would like to thank the anonymous reviewers for their insightful feedback and constructive suggestions, which have significantly improved the quality of this paper. 
This work was supported by the National Natural Science Foundation of China (NSFC) under Grant Nos. 62125201, U24B20174, and U25B6003.

\section*{Ethics Statement}

This research aims to advance the field of artificial intelligence, particularly in the area of Unified Multimodal Models. We recognize that, like other powerful generative models, the technologies proposed in this study also carry potential risks of misuse, such as the creation of misinformation, biased, or harmful content. Our primary objective is to explore architectural efficiency to build more powerful and scalable models, and we believe this will make a valuable contribution to science.

The datasets used to train and fine-tune our models are primarily publicly available and widely used benchmark datasets in the academic community. For any internally collected data, we have ensured that its acquisition and processing adhere to principles of responsibility. We have not specifically filtered web-based datasets for bias, and therefore, the model may reflect social biases present in the data. We encourage responsible downstream use and further research into mitigating the potential negative impacts of generative models. Our work is intended solely for research purposes and is shared with the community to foster innovation and deepen understanding.

\section*{Reproducibility Statement}

We are committed to ensuring the full reproducibility of our work. To facilitate verification and future extensions, we provide detailed descriptions of the model architecture, training setup, datasets, evaluation protocols, and implementation details throughout the main paper and appendix.

\paragraph{Code.} 
The complete source code for the Uni-X architecture, training pipeline, and evaluation scripts is publicly available at: \url{https://github.com/CURRENTF/Uni-X}.
The repository includes model definitions, data preprocessing utilities, configuration files, and instructions for reproducing both ablation and scaling experiments.

\paragraph{Architecture and Implementation.} 
The description of the Uni-X architecture is provided in Section \ref{sect:method:arch}. 
%%%
Implementation details, including the base models (Qwen2.5-1.5B and Qwen2.5-3B), the VQGAN tokenizer configuration, training framework (Liquid and HuggingFace Transformers), optimization tools (FlashAttention 2 and DeepSpeed ZeRO2), and image generation settings (e.g., CFG = 4.0), are documented in Section \ref{sect:expt:setup}. 
%%%
Adaptations made to baseline methods under the VQ+AR setting are fully described in Appendix \ref{appendix:baseline_impl} to ensure fair and transparent comparison.

\paragraph{Datasets and Evaluation Benchmarks.} 
All pre-training datasets (including text and multimodal corpora) and supervised fine-tuning datasets are listed in Section \ref{sect:expt:setup}, along with token statistics.
Evaluation benchmarks for text-only tasks, image generation, multimodal understanding, and image editing are also described in Section \ref{sect:expt:setup} and corresponding result tables.
%%%
Where prompt rewriting or evaluation-specific adjustments were applied (e.g., GenEval prompt expansion), these procedures are explicitly documented.

\paragraph{Hyperparameters and Training Settings.} 
Key training hyperparameters for ablation experiments, including learning rate, warmup ratio, batch size, and scheduler details, are specified in Section \ref{sect:expt:setup}. 
Scaling experiments and total token counts (e.g., 140B / 240B settings) are also clearly reported in Section \ref{sect:expt:analysis}.
%%%
These details ensure that experimental comparisons are conducted under fully specified and reproducible conditions.

We believe that the combination of open-source code, precise architectural definitions, transparent dataset descriptions, and explicitly reported training hyperparameters enables the research community to faithfully reproduce our results and build upon the Uni-X framework for future unified multimodal modeling research.

% \section*{The Use of Large Language Models (LLMs)}
% LLMs are used mainly as auxiliary tools to aid and polish the writing of the paper. 

\bibliography{reference}

@article{meng2022locating,
  title={Locating and editing factual associations in gpt},
  author={Meng, Kevin and Bau, David and Andonian, Alex and Belinkov, Yonatan},
  journal={Advances in neural information processing systems},
  volume={35},
  pages={17359--17372},
  year={2022},
  url={https://openreview.net/forum?id=-h6WAS6eE4}
}

@misc{geva_transformer_2021,
	title = {Transformer {Feed}-{Forward} {Layers} {Are} {Key}-{Value} {Memories}},
	url = {http://arxiv.org/abs/2012.14913},
	language = {en},
	urldate = {2023-07-12},
	publisher = {arXiv},
	author = {Geva, Mor and Schuster, Roei and Berant, Jonathan and Levy, Omer},
	month = sep,
	year = {2021},
	note = {arXiv:2012.14913 [cs]},
}

@misc{li_snapkv_2024,
	title = {{SnapKV}: {LLM} {Knows} {What} {You} are {Looking} for {Before} {Generation}},
	shorttitle = {{SnapKV}},
	url = {http://arxiv.org/abs/2404.14469},
	doi = {10.48550/arXiv.2404.14469},
	urldate = {2025-05-20},
	publisher = {arXiv},
	author = {Li, Yuhong and Huang, Yingbing and Yang, Bowen and Venkitesh, Bharat and Locatelli, Acyr and Ye, Hanchen and Cai, Tianle and Lewis, Patrick and Chen, Deming},
	month = jun,
	year = {2024},
	note = {arXiv:2404.14469 [cs]}
}

@misc{wu_liquid_2025,
	title = {Liquid: {Language} {Models} are {Scalable} and {Unified} {Multi}-modal {Generators}},
	shorttitle = {Liquid},
	url = {http://arxiv.org/abs/2412.04332},
	doi = {10.48550/arXiv.2412.04332},
	language = {en},
	urldate = {2025-06-12},
	publisher = {arXiv},
	author = {Wu, Junfeng and Jiang, Yi and Ma, Chuofan and Liu, Yuliang and Zhao, Hengshuang and Yuan, Zehuan and Bai, Song and Bai, Xiang},
	month = apr,
	year = {2025},
}

@misc{deng_emerging_2025,
	title = {Emerging {Properties} in {Unified} {Multimodal} {Pretraining}},
	copyright = {Creative Commons Attribution 4.0 International},
	url = {https://arxiv.org/abs/2505.14683},
	doi = {10.48550/ARXIV.2505.14683},
	language = {en},
	urldate = {2025-06-13},
	publisher = {arXiv},
	author = {Deng, Chaorui and Zhu, Deyao and Li, Kunchang and Gou, Chenhui and Li, Feng and Wang, Zeyu and Zhong, Shu and Yu, Weihao and Nie, Xiaonan and Song, Ziang and Shi, Guang and Fan, Haoqi},
	year = {2025},
}

@misc{mmlu_hendrycks2021measuringmassivemultitasklanguage,
      title={Measuring Massive Multitask Language Understanding}, 
      author={Dan Hendrycks and Collin Burns and Steven Basart and Andy Zou and Mantas Mazeika and Dawn Song and Jacob Steinhardt},
      year={2021},
      eprint={2009.03300},
      archivePrefix={arXiv},
      primaryClass={cs.CY},
      url={https://arxiv.org/abs/2009.03300}, 
}

@misc{arc_clark2018thinksolvedquestionanswering,
      title={Think you have Solved Question Answering? Try ARC, the AI2 Reasoning Challenge}, 
      author={Peter Clark and Isaac Cowhey and Oren Etzioni and Tushar Khot and Ashish Sabharwal and Carissa Schoenick and Oyvind Tafjord},
      year={2018},
      eprint={1803.05457},
      archivePrefix={arXiv},
      primaryClass={cs.AI},
      url={https://arxiv.org/abs/1803.05457}, 
}

@misc{chen_janus-pro_2025,
	title = {Janus-{Pro}: {Unified} {Multimodal} {Understanding} and {Generation} with {Data} and {Model} {Scaling}},
	shorttitle = {Janus-{Pro}},
	url = {http://arxiv.org/abs/2501.17811},
	doi = {10.48550/arXiv.2501.17811},
	language = {en},
	urldate = {2025-07-15},
	publisher = {arXiv},
	author = {Chen, Xiaokang and Wu, Zhiyu and Liu, Xingchao and Pan, Zizheng and Liu, Wen and Xie, Zhenda and Yu, Xingkai and Ruan, Chong},
	month = jan,
	year = {2025},
}

@misc{wu_janus_2024,
	title = {Janus: {Decoupling} {Visual} {Encoding} for {Unified} {Multimodal} {Understanding} and {Generation}},
	shorttitle = {Janus},
	url = {http://arxiv.org/abs/2410.13848},
	doi = {10.48550/arXiv.2410.13848},
	language = {en},
	urldate = {2025-07-15},
	publisher = {arXiv},
	author = {Wu, Chengyue and Chen, Xiaokang and Wu, Zhiyu and Ma, Yiyang and Liu, Xingchao and Pan, Zizheng and Liu, Wen and Xie, Zhenda and Yu, Xingkai and Ruan, Chong and Luo, Ping},
	month = oct,
	year = {2024},
}

@misc{team_chameleon_2025,
	title = {Chameleon: {Mixed}-{Modal} {Early}-{Fusion} {Foundation} {Models}},
	shorttitle = {Chameleon},
	url = {http://arxiv.org/abs/2405.09818},
	doi = {10.48550/arXiv.2405.09818},
	language = {en},
	urldate = {2025-07-21},
	publisher = {arXiv},
	author = {Team, Chameleon},
	month = mar,
	year = {2025},
	note = {arXiv:2405.09818 [cs]},
}

@misc{chen_blip3-o_2025,
	title = {{BLIP3}-o: {A} {Family} of {Fully} {Open} {Unified} {Multimodal} {Models}-{Architecture}, {Training} and {Dataset}},
	shorttitle = {{BLIP3}-o},
	url = {http://arxiv.org/abs/2505.09568},
	doi = {10.48550/arXiv.2505.09568},
	language = {en},
	urldate = {2025-07-22},
	publisher = {arXiv},
	author = {Chen, Jiuhai and Xu, Zhiyang and Pan, Xichen and Hu, Yushi and Qin, Can and Goldstein, Tom and Huang, Lifu and Zhou, Tianyi and Xie, Saining and Savarese, Silvio and Xue, Le and Xiong, Caiming and Xu, Ran},
	month = may,
	year = {2025},
}

@misc{li_unifork_2025,
	title = {{UniFork}: {Exploring} {Modality} {Alignment} for {Unified} {Multimodal} {Understanding} and {Generation}},
	shorttitle = {{UniFork}},
	url = {http://arxiv.org/abs/2506.17202},
	doi = {10.48550/arXiv.2506.17202},
	language = {en},
	urldate = {2025-07-31},
	publisher = {arXiv},
	author = {Li, Teng and Lu, Quanfeng and Zhao, Lirui and Li, Hao and Zhu, Xizhou and Qiao, Yu and Zhang, Jun and Shao, Wenqi},
	month = jun,
	year = {2025},
}

@misc{shi_lmfusion_2025,
	title = {{LMFusion}: {Adapting} {Pretrained} {Language} {Models} for {Multimodal} {Generation}},
	shorttitle = {{LMFusion}},
	url = {http://arxiv.org/abs/2412.15188},
	doi = {10.48550/arXiv.2412.15188},
	urldate = {2025-08-11},
	publisher = {arXiv},
	author = {Shi, Weijia and Han, Xiaochuang and Zhou, Chunting and Liang, Weixin and Lin, Xi Victoria and Zettlemoyer, Luke and Yu, Lili},
	month = feb,
	year = {2025},
}

@misc{wu_omnigen2_2025,
	title = {{OmniGen2}: {Exploration} to {Advanced} {Multimodal} {Generation}},
	shorttitle = {{OmniGen2}},
	url = {http://arxiv.org/abs/2506.18871},
	doi = {10.48550/arXiv.2506.18871},
	urldate = {2025-08-11},
	publisher = {arXiv},
	author = {Wu, Chenyuan and Zheng, Pengfei and Yan, Ruiran and Xiao, Shitao and Luo, Xin and Wang, Yueze and Li, Wanli and Jiang, Xiyan and Liu, Yexin and Zhou, Junjie and Liu, Ze and Xia, Ziyi and Li, Chaofan and Deng, Haoge and Wang, Jiahao and Luo, Kun and Zhang, Bo and Lian, Defu and Wang, Xinlong and Wang, Zhongyuan and Huang, Tiejun and Liu, Zheng},
	month = jun,
	year = {2025},
}

@misc{xie_show-o2_2025,
	title = {Show-o2: {Improved} {Native} {Unified} {Multimodal} {Models}},
	shorttitle = {Show-o2},
	url = {http://arxiv.org/abs/2506.15564},
	doi = {10.48550/arXiv.2506.15564},
	urldate = {2025-08-13},
	publisher = {arXiv},
	author = {Xie, Jinheng and Yang, Zhenheng and Shou, Mike Zheng},
	month = jun,
	year = {2025},
}

@misc{liao_mogao_2025,
	title = {Mogao: {An} {Omni} {Foundation} {Model} for {Interleaved} {Multi}-{Modal} {Generation}},
	shorttitle = {Mogao},
	url = {http://arxiv.org/abs/2505.05472},
	doi = {10.48550/arXiv.2505.05472},
	urldate = {2025-08-13},
	publisher = {arXiv},
	author = {Liao, Chao and Liu, Liyang and Wang, Xun and Luo, Zhengxiong and Zhang, Xinyu and Zhao, Wenliang and Wu, Jie and Li, Liang and Tian, Zhi and Huang, Weilin},
	month = may,
	year = {2025},
}

@misc{geng_x-omni_2025,
	title = {X-{Omni}: {Reinforcement} {Learning} {Makes} {Discrete} {Autoregressive} {Image} {Generative} {Models} {Great} {Again}},
	shorttitle = {X-{Omni}},
	url = {http://arxiv.org/abs/2507.22058},
	doi = {10.48550/arXiv.2507.22058},
	urldate = {2025-08-17},
	publisher = {arXiv},
	author = {Geng, Zigang and Wang, Yibing and Ma, Yeyao and Li, Chen and Rao, Yongming and Gu, Shuyang and Zhong, Zhao and Lu, Qinglin and Hu, Han and Zhang, Xiaosong and Linus and Wang, Di and Jiang, Jie},
	month = jul,
	year = {2025},
}

@misc{team_nextstep-1_2025,
	title = {{NextStep}-1: {Toward} {Autoregressive} {Image} {Generation} with {Continuous} {Tokens} at {Scale}},
	shorttitle = {{NextStep}-1},
	url = {http://arxiv.org/abs/2508.10711},
	doi = {10.48550/arXiv.2508.10711},
	urldate = {2025-08-18},
	publisher = {arXiv},
	author = {Team, NextStep and Han, Chunrui and Li, Guopeng and Wu, Jingwei and Sun, Quan and Cai, Yan and Peng, Yuang and Ge, Zheng and Zhou, Deyu and Tang, Haomiao and Zhou, Hongyu and Liu, Kenkun and Huang, Ailin and Wang, Bin and Miao, Changxin and Sun, Deshan and Yu, En and Yin, Fukun and Yu, Gang and Nie, Hao and Lv, Haoran and Hu, Hanpeng and Wang, Jia and Zhou, Jian and Sun, Jianjian and Tan, Kaijun and An, Kang and Lin, Kangheng and Zhao, Liang and Chen, Mei and Xing, Peng and Wang, Rui and Liu, Shiyu and Xia, Shutao and You, Tianhao and Ji, Wei and Zeng, Xianfang and Han, Xin and Zhang, Xuelin and Wei, Yana and Xu, Yanming and Jiang, Yimin and Wang, Yingming and Zhou, Yu and Han, Yucheng and Meng, Ziyang and Jiao, Binxing and Jiang, Daxin and Zhang, Xiangyu and Zhu, Yibo},
	month = aug,
	year = {2025},
}

@misc{li_mini-gemini_2024,
	title = {Mini-{Gemini}: {Mining} the {Potential} of {Multi}-modality {Vision} {Language} {Models}},
	shorttitle = {Mini-{Gemini}},
	url = {http://arxiv.org/abs/2403.18814},
	doi = {10.48550/arXiv.2403.18814},
	urldate = {2025-09-02},
	publisher = {arXiv},
	author = {Li, Yanwei and Zhang, Yuechen and Wang, Chengyao and Zhong, Zhisheng and Chen, Yixin and Chu, Ruihang and Liu, Shaoteng and Jia, Jiaya},
	month = mar,
	year = {2024},
}

@misc{jin_unified_2024,
	title = {Unified {Language}-{Vision} {Pretraining} in {LLM} with {Dynamic} {Discrete} {Visual} {Tokenization}},
	url = {http://arxiv.org/abs/2309.04669},
	doi = {10.48550/arXiv.2309.04669},
	language = {en},
	urldate = {2025-09-06},
	publisher = {arXiv},
	author = {Jin, Yang and Xu, Kun and Xu, Kun and Chen, Liwei and Liao, Chao and Tan, Jianchao and Huang, Quzhe and Chen, Bin and Lei, Chenyi and Liu, An and Song, Chengru and Lei, Xiaoqiang and Zhang, Di and Ou, Wenwu and Gai, Kun and Mu, Yadong},
	month = mar,
	year = {2024},
}

@article{zhao2024monoformer,
  title={Monoformer: One transformer for both diffusion and autoregression},
  author={Zhao, Chuyang and Song, Yuxing and Wang, Wenhao and Feng, Haocheng and Ding, Errui and Sun, Yifan and Xiao, Xinyan and Wang, Jingdong},
  journal={arXiv preprint arXiv:2409.16280},
  year={2024},
  url={https://arxiv.org/abs/2409.16280}
}

@misc{ge2025seedxmultimodalmodelsunified,
      title={SEED-X: Multimodal Models with Unified Multi-granularity Comprehension and Generation}, 
      author={Yuying Ge and Sijie Zhao and Jinguo Zhu and Yixiao Ge and Kun Yi and Lin Song and Chen Li and Xiaohan Ding and Ying Shan},
      year={2025},
      eprint={2404.14396},
      archivePrefix={arXiv},
      primaryClass={cs.CV},
      url={https://arxiv.org/abs/2404.14396}, 
}

@String(AAAI = {AAAI})

@inproceedings{CLIP,
  title={Learning transferable visual models from natural language supervision},
  author={Radford, Alec and Kim, Jong Wook and Hallacy, Chris and Ramesh, Aditya and Goh, Gabriel and Agarwal, Sandhini and Sastry, Girish and Askell, Amanda and Mishkin, Pamela and Clark, Jack and others},
  booktitle={International conference on machine learning},
  pages={8748--8763},
  year={2021},
  organization={PMLR},
  url={https://proceedings.mlr.press/v139/radford21a/radford21a.pdf}
}

@article{llama,
  title={Llama: Open and efficient foundation language models},
  author={Touvron, Hugo and Lavril, Thibaut and Izacard, Gautier and Martinet, Xavier and Lachaux, Marie-Anne and Lacroix, Timoth{\'e}e and Rozi{\`e}re, Baptiste and Goyal, Naman and Hambro, Eric and Azhar, Faisal and others},
  journal={arXiv preprint arXiv:2302.13971},
  year={2023},
  url={https://arxiv.org/abs/2302.13971}
}

@inproceedings{gpt3,
  title={Language models are few-shot learners},
  author={Brown, Tom and Mann, Benjamin and Ryder, Nick and Subbiah, Melanie and Kaplan, Jared D and Dhariwal, Prafulla and Neelakantan, Arvind and Shyam, Pranav and Sastry, Girish and Askell, Amanda and others},
  booktitle={Proceedings of the 34th International Conference on Neural Information Processing Systems},
  volume={33},
  pages={1877--1901},
  year={2020},
  url={https://proceedings.neurips.cc/paper_files/paper/2020/file/1457c0d6bfcb4967418bfb8ac142f64a-Paper.pdf}
}

@article{team2024gemma,
  title={Gemma: Open models based on gemini research and technology},
  author={Team, Gemma and Mesnard, Thomas and Hardin, Cassidy and Dadashi, Robert and Bhupatiraju, Surya and Pathak, Shreya and Sifre, Laurent and Rivi{\`e}re, Morgane and Kale, Mihir Sanjay and Love, Juliette and others},
  journal={arXiv preprint arXiv:2403.08295},
  year={2024},
  url={https://arxiv.org/abs/2403.08295}
}

@article{team2024gemma2,
  title={Gemma 2: Improving open language models at a practical size},
  author={Team, Gemma and Riviere, Morgane and Pathak, Shreya and Sessa, Pier Giuseppe and Hardin, Cassidy and Bhupatiraju, Surya and Hussenot, L{\'e}onard and Mesnard, Thomas and Shahriari, Bobak and Ram{\'e}, Alexandre and others},
  journal={arXiv preprint arXiv:2408.00118},
  year={2024},
  url={https://arxiv.org/abs/2408.00118}
}

@article{llava,
  title={Visual instruction tuning},
  author={Liu, Haotian and Li, Chunyuan and Wu, Qingyang and Lee, Yong Jae},
  journal={Advances in neural information processing systems},
  volume={36},
  year={2024},
  url={https://openreview.net/forum?id=w0H2xGHlkw}
}

@article{internvl1.5,
  title={How far are we to gpt-4v? closing the gap to commercial multimodal models with open-source suites},
  author={Chen, Zhe and Wang, Weiyun and Tian, Hao and Ye, Shenglong and Gao, Zhangwei and Cui, Erfei and Tong, Wenwen and Hu, Kongzhi and Luo, Jiapeng and Ma, Zheng and others},
  journal={arXiv preprint arXiv:2404.16821},
  year={2024},
  url={https://arxiv.org/abs/2404.16821}
}

@article{chen2023sharegpt4v,
  title={Sharegpt4v: Improving large multi-modal models with better captions},
  author={Chen, Lin and Li, Jinsong and Dong, Xiaoyi and Zhang, Pan and He, Conghui and Wang, Jiaqi and Zhao, Feng and Lin, Dahua},
  journal={arXiv preprint arXiv:2311.12793},
  year={2023},
  url={https://arxiv.org/abs/2311.12793}
}

@misc{liu2024llavanext,
    title={LLaVA-NeXT: Improved reasoning, OCR, and world knowledge},
    url={https://llava-vl.github.io/blog/2024-01-30-llava-next/},
    author={Liu, Haotian and Li, Chunyuan and Li, Yuheng and Li, Bo and Zhang, Yuanhan and Shen, Sheng and Lee, Yong Jae},
    month={January},
    year={2024}
}

@article{liu2024world_lwm,
  title={World model on million-length video and language with ringattention},
  author={Liu, Hao and Yan, Wilson and Zaharia, Matei and Abbeel, Pieter},
  journal={arXiv preprint arXiv:2402.08268},
  year={2024},
  url={https://arxiv.org/abs/2402.08268}
}

@article{xie2024show,
  title={Show-o: One single transformer to unify multimodal understanding and generation},
  author={Xie, Jinheng and Mao, Weijia and Bai, Zechen and Zhang, David Junhao and Wang, Weihao and Lin, Kevin Qinghong and Gu, Yuchao and Chen, Zhijie and Yang, Zhenheng and Shou, Mike Zheng},
  journal={arXiv preprint arXiv:2408.12528},
  year={2024},
  url={https://arxiv.org/abs/2408.12528}
}

@article{wu2024vila,
  title={Vila-u: a unified foundation model integrating visual understanding and generation},
  author={Wu, Yecheng and Zhang, Zhuoyang and Chen, Junyu and Tang, Haotian and Li, Dacheng and Fang, Yunhao and Zhu, Ligeng and Xie, Enze and Yin, Hongxu and Yi, Li and others},
  journal={arXiv preprint arXiv:2409.04429},
  year={2024},
  url={https://arxiv.org/abs/2409.04429}
}

@article{wang2024emu3,
  title={Emu3: Next-token prediction is all you need},
  author={Wang, Xinlong and Zhang, Xiaosong and Luo, Zhengxiong and Sun, Quan and Cui, Yufeng and Wang, Jinsheng and Zhang, Fan and Wang, Yueze and Li, Zhen and Yu, Qiying and others},
  journal={arXiv preprint arXiv:2409.18869},
  year={2024},
  url={https://arxiv.org/abs/2409.18869}
}

@article{dreamllm,
  title={Dreamllm: Synergistic multimodal comprehension and creation},
  author={Dong, Runpei and Han, Chunrui and Peng, Yuang and Qi, Zekun and Ge, Zheng and Yang, Jinrong and Zhao, Liang and Sun, Jianjian and Zhou, Hongyu and Wei, Haoran and others},
  journal={arXiv preprint arXiv:2309.11499},
  year={2023},
  url={https://arxiv.org/abs/2309.11499}
}

@inproceedings{esser2021taming,
  title={Taming transformers for high-resolution image synthesis},
  author={Esser, Patrick and Rombach, Robin and Ommer, Bjorn},
  booktitle={Proceedings of the IEEE/CVF conference on computer vision and pattern recognition},
  pages={12873--12883},
  year={2021},
  url={https://openaccess.thecvf.com/content/CVPR2021/html/Esser_Taming_Transformers_for_High-Resolution_Image_Synthesis_CVPR_2021_paper.html}
}

@misc{pan2023journeydb,
      title={JourneyDB: A Benchmark for Generative Image Understanding}, 
      author={Junting Pan and Keqiang Sun and Yuying Ge and Hao Li and Haodong Duan and Xiaoshi Wu and Renrui Zhang and Aojun Zhou and Zipeng Qin and Yi Wang and Jifeng Dai and Yu Qiao and Hongsheng Li},
      year={2023},
      eprint={2307.00716},
      archivePrefix={arXiv},
      primaryClass={cs.CV},
      url={https://arxiv.org/abs/2307.00716}
}

@article{POPE,
  title={Evaluating object hallucination in large vision-language models},
  author={Li, Yifan and Du, Yifan and Zhou, Kun and Wang, Jinpeng and Zhao, Wayne Xin and Wen, Ji-Rong},
  journal={arXiv preprint arXiv:2305.10355},
  year={2023},
  url={https://arxiv.org/abs/2305.10355}
}

@misc{mme,
      title={MME: A Comprehensive Evaluation Benchmark for Multimodal Large Language Models}, 
      author={Chaoyou Fu and Peixian Chen and Yunhang Shen and Yulei Qin and Mengdan Zhang and Xu Lin and Jinrui Yang and Xiawu Zheng and Ke Li and Xing Sun and Yunsheng Wu and Rongrong Ji},
      year={2024},
      eprint={2306.13394},
      archivePrefix={arXiv},
      primaryClass={cs.CV},
      url={https://arxiv.org/abs/2306.13394}
}

@article{seedbench,
  title={Seed-bench: Benchmarking multimodal llms with generative comprehension},
  author={Li, Bohao and Wang, Rui and Wang, Guangzhi and Ge, Yuying and Ge, Yixiao and Shan, Ying},
  journal={arXiv preprint arXiv:2307.16125},
  year={2023},
  url={https://arxiv.org/abs/2307.16125}
}

@misc{liu2024mmbenchmultimodalmodelallaround,
      title={MMBench: Is Your Multi-modal Model an All-around Player?}, 
      author={Yuan Liu and Haodong Duan and Yuanhan Zhang and Bo Li and Songyang Zhang and Wangbo Zhao and Yike Yuan and Jiaqi Wang and Conghui He and Ziwei Liu and Kai Chen and Dahua Lin},
      year={2024},
      eprint={2307.06281},
      archivePrefix={arXiv},
      primaryClass={cs.CV},
      url={https://arxiv.org/abs/2307.06281}, 
}

@article{transfusion,
  title={Transfusion: Predict the next token and diffuse images with one multi-modal model},
  author={Zhou, Chunting and Yu, Lili and Babu, Arun and Tirumala, Kushal and Yasunaga, Michihiro and Shamis, Leonid and Kahn, Jacob and Ma, Xuezhe and Zettlemoyer, Luke and Levy, Omer},
  journal={arXiv preprint arXiv:2408.11039},
  year={2024},
  url={https://arxiv.org/abs/2408.11039}
}

@misc{sun2025transformerlayerspainters,
      title={Transformer Layers as Painters}, 
      author={Qi Sun and Marc Pickett and Aakash Kumar Nain and Llion Jones},
      year={2025},
      eprint={2407.09298},
      archivePrefix={arXiv},
      primaryClass={cs.CL},
      url={https://arxiv.org/abs/2407.09298}, 
}

@inproceedings{dao2022flashattention,
  title={Flashattention: Fast and memory-efficient exact attention with io-awareness},
  author={Dao, Tri and Fu, Dan and Ermon, Stefano and Rudra, Atri and R{\'e}, Christopher},
  booktitle={Proceedings of the 36th International Conference on Neural Information Processing Systems (NeurIPS)},
  pages={16344--16359},
  year={2022},
  url={https://arxiv.org/abs/2205.14135}
}

@article{wolf2019huggingface,
  title={{Huggingface's} transformers: State-of-the-art natural language processing},
  author={Thomas Wolf and Lysandre Debut and Victor Sanh and Julien Chaumond and Clement Delangue and Anthony Moi and Pierric Cistac and Tim Rault and Rémi Louf and Morgan Funtowicz and Joe Davison and Sam Shleifer and Patrick von Platen and Clara Ma and Yacine Jernite and Julien Plu and Canwen Xu and Teven Le Scao and Sylvain Gugger and Mariama Drame and Quentin Lhoest and Alexander M. Rush},
  journal={arXiv preprint arXiv:1910.03771},
  year={2019},
  url={https://arxiv.org/abs/1910.03771}
}

@inproceedings{aminabadi2022deepspeed,
  title={{DeepSpeed-inference}: enabling efficient inference of transformer models at unprecedented scale},
  author={Aminabadi, Reza Yazdani and Rajbhandari, Samyam and Awan, Ammar Ahmad and Li, Cheng and Li, Du and Zheng, Elton and Ruwase, Olatunji and Smith, Shaden and Zhang, Minjia and Rasley, Jeff and others},
  booktitle={Proceedings of the International Conference on High Performance Computing, Networking, Storage and Analysis (SC'22)},
  pages={1--15},
  year={2022},
  url={https://dl.acm.org/doi/abs/10.5555/3571885.3571946}
}

@inproceedings{clark2019boolq,
  title={{BoolQ}: Exploring the Surprising Difficulty of Natural Yes/No Questions},
  author={Clark, Christopher and Lee, Kenton and Chang, Ming-Wei and Kwiatkowski, Tom and Collins, Michael and Toutanova, Kristina},
  booktitle={Proceedings of the 2019 Conference of the North American Chapter of the Association for Computational Linguistics: Human Language Technologies (NAACL-HLT)},
  pages={2924--2936},
  year={2019},
  url={https://aclanthology.org/N19-1300/}
}

@inproceedings{sakaguchi2020winogrande,
  title={{WinoGrande}: An Adversarial Winograd Schema Challenge at Scale},
  author={Sakaguchi, Keisuke and Le Bras, Ronan and Bhagavatula, Chandra and Choi, Yejin},
  booktitle={Proceedings of the AAAI Conference on Artificial Intelligence (AAAI)},
  pages={8732--8740},
  year={2020},
  url={https://ojs.aaai.org/index.php/AAAI/article/view/6399}
}

@inproceedings{penedo2024fineweb,
  title={The {FineWeb} Datasets: Decanting the Web for the Finest Text Data at Scale},
  author={Penedo, Guilherme and Kydl{\'\i}{\v{c}}ek, Hynek and Lozhkov, Anton and Mitchell, Margaret and Raffel, Colin A and Von Werra, Leandro and Wolf, Thomas and others},
  booktitle={Proceedings of the 38th International Conference on Neural Information Processing Systems (NeurIPS)},
  pages={30811--30849},
  year={2024},
  url={https://openreview.net/forum?id=n6SCkn2QaG}
}

@inproceedings{geva2021transformer,
  title={Transformer Feed-Forward Layers Are Key-Value Memories},
  author={Geva, Mor and Schuster, Roei and Berant, Jonathan and Levy, Omer},
  booktitle={Proceedings of the 2021 Conference on Empirical Methods in Natural Language Processing (EMNLP)},
  pages={5484--5495},
  year={2021},
  url={https://aclanthology.org/2021.emnlp-main.446/},
}

@article{achiam2023gpt4,
  title={{GPT-4} {T}echnical {R}eport},
  author={Achiam, Josh and Adler, Steven and Agarwal, Sandhini and Ahmad, Lama and Akkaya, Ilge and Aleman, Florencia Leoni and Almeida, Diogo and Altenschmidt, Janko and Altman, Sam and Anadkat, Shyamal and others},
  journal={arXiv preprint arXiv:2303.08774},
  year={2023},
  url={https://arxiv.org/abs/2303.08774}
}

@article{yang2024qwen2_5,
  title={{Qwen2.5} {T}echnical {R}eport},
  author={Yang, An and Yang, Baosong and Zhang, Beichen and Hui, Binyuan and Zheng, Bo and Yu, Bowen and Li, Chengyuan and Liu, Dayiheng and Huang, Fei and Wei, Haoran and others},
  journal={arXiv preprint arXiv:2412.15115},
  year={2024},
  url={https://arxiv.org/abs/2412.15115}
}

@misc{wang2024cci30hqlargescalechinesedataset,
      title={CCI3.0-HQ: a large-scale Chinese dataset of high quality designed for pre-training large language models}, 
      author={Liangdong Wang and Bo-Wen Zhang and Chengwei Wu and Hanyu Zhao and Xiaofeng Shi and Shuhao Gu and Jijie Li and Quanyue Ma and TengFei Pan and Guang Liu},
      year={2024},
      eprint={2410.18505},
      archivePrefix={arXiv},
      primaryClass={cs.CL},
      url={https://arxiv.org/abs/2410.18505}, 
}

@misc{li2025datacomplmsearchgenerationtraining,
      title={DataComp-LM: In search of the next generation of training sets for language models}, 
      author={Jeffrey Li and Alex Fang and Georgios Smyrnis and Maor Ivgi and Matt Jordan and Samir Gadre and Hritik Bansal and Etash Guha and Sedrick Keh and Kushal Arora and Saurabh Garg and Rui Xin and Niklas Muennighoff and Reinhard Heckel and Jean Mercat and Mayee Chen and Suchin Gururangan and Mitchell Wortsman and Alon Albalak and Yonatan Bitton and Marianna Nezhurina and Amro Abbas and Cheng-Yu Hsieh and Dhruba Ghosh and Josh Gardner and Maciej Kilian and Hanlin Zhang and Rulin Shao and Sarah Pratt and Sunny Sanyal and Gabriel Ilharco and Giannis Daras and Kalyani Marathe and Aaron Gokaslan and Jieyu Zhang and Khyathi Chandu and Thao Nguyen and Igor Vasiljevic and Sham Kakade and Shuran Song and Sujay Sanghavi and Fartash Faghri and Sewoong Oh and Luke Zettlemoyer and Kyle Lo and Alaaeldin El-Nouby and Hadi Pouransari and Alexander Toshev and Stephanie Wang and Dirk Groeneveld and Luca Soldaini and Pang Wei Koh and Jenia Jitsev and Thomas Kollar and Alexandros G. Dimakis and Yair Carmon and Achal Dave and Ludwig Schmidt and Vaishaal Shankar},
      year={2025},
      eprint={2406.11794},
      archivePrefix={arXiv},
      primaryClass={cs.LG},
      url={https://arxiv.org/abs/2406.11794}, 
}

@misc{li2023starcodersourceyou,
      title={StarCoder: may the source be with you!}, 
      author={Raymond Li and Loubna Ben Allal and Yangtian Zi and Niklas Muennighoff and Denis Kocetkov and Chenghao Mou and Marc Marone and Christopher Akiki and Jia Li and Jenny Chim and Qian Liu and Evgenii Zheltonozhskii and Terry Yue Zhuo and Thomas Wang and Olivier Dehaene and Mishig Davaadorj and Joel Lamy-Poirier and João Monteiro and Oleh Shliazhko and Nicolas Gontier and Nicholas Meade and Armel Zebaze and Ming-Ho Yee and Logesh Kumar Umapathi and Jian Zhu and Benjamin Lipkin and Muhtasham Oblokulov and Zhiruo Wang and Rudra Murthy and Jason Stillerman and Siva Sankalp Patel and Dmitry Abulkhanov and Marco Zocca and Manan Dey and Zhihan Zhang and Nour Fahmy and Urvashi Bhattacharyya and Wenhao Yu and Swayam Singh and Sasha Luccioni and Paulo Villegas and Maxim Kunakov and Fedor Zhdanov and Manuel Romero and Tony Lee and Nadav Timor and Jennifer Ding and Claire Schlesinger and Hailey Schoelkopf and Jan Ebert and Tri Dao and Mayank Mishra and Alex Gu and Jennifer Robinson and Carolyn Jane Anderson and Brendan Dolan-Gavitt and Danish Contractor and Siva Reddy and Daniel Fried and Dzmitry Bahdanau and Yacine Jernite and Carlos Muñoz Ferrandis and Sean Hughes and Thomas Wolf and Arjun Guha and Leandro von Werra and Harm de Vries},
      year={2023},
      eprint={2305.06161},
      archivePrefix={arXiv},
      primaryClass={cs.CL},
      url={https://arxiv.org/abs/2305.06161}, 
}

@misc{russakovsky2015imagenetlargescalevisual,
      title={ImageNet Large Scale Visual Recognition Challenge}, 
      author={Olga Russakovsky and Jia Deng and Hao Su and Jonathan Krause and Sanjeev Satheesh and Sean Ma and Zhiheng Huang and Andrej Karpathy and Aditya Khosla and Michael Bernstein and Alexander C. Berg and Li Fei-Fei},
      year={2015},
      eprint={1409.0575},
      archivePrefix={arXiv},
      primaryClass={cs.CV},
      url={https://arxiv.org/abs/1409.0575}, 
}

@misc{huggingface2025finevision,
  author       = {HuggingFaceM4},
  title        = {FineVision Dataset},
  year         = {2025},
  publisher    = {Hugging Face},
  howpublished = {\url{https://huggingface.co/datasets/HuggingFaceM4/FineVision}}
}

@misc{ghosh2023genevalobjectfocusedframeworkevaluating,
      title={GenEval: An Object-Focused Framework for Evaluating Text-to-Image Alignment}, 
      author={Dhruba Ghosh and Hanna Hajishirzi and Ludwig Schmidt},
      year={2023},
      eprint={2310.11513},
      archivePrefix={arXiv},
      primaryClass={cs.CV},
      url={https://arxiv.org/abs/2310.11513}, 
}

@misc{dpg_bench,
      title={ELLA: Equip Diffusion Models with LLM for Enhanced Semantic Alignment}, 
      author={Xiwei Hu and Rui Wang and Yixiao Fang and Bin Fu and Pei Cheng and Gang Yu},
      year={2024},
      eprint={2403.05135},
      archivePrefix={arXiv},
      primaryClass={cs.CV},
      url={https://arxiv.org/abs/2403.05135}, 
}

@misc{openorca,
      title={Orca: Progressive Learning from Complex Explanation Traces of GPT-4}, 
      author={Subhabrata Mukherjee and Arindam Mitra and Ganesh Jawahar and Sahaj Agarwal and Hamid Palangi and Ahmed Awadallah},
      year={2023},
      eprint={2306.02707},
      archivePrefix={arXiv},
      primaryClass={cs.CL},
      url={https://arxiv.org/abs/2306.02707}, 
}

@misc{tschannen2025siglip2multilingualvisionlanguage,
      title={SigLIP 2: Multilingual Vision-Language Encoders with Improved Semantic Understanding, Localization, and Dense Features}, 
      author={Michael Tschannen and Alexey Gritsenko and Xiao Wang and Muhammad Ferjad Naeem and Ibrahim Alabdulmohsin and Nikhil Parthasarathy and Talfan Evans and Lucas Beyer and Ye Xia and Basil Mustafa and Olivier Hénaff and Jeremiah Harmsen and Andreas Steiner and Xiaohua Zhai},
      year={2025},
      eprint={2502.14786},
      archivePrefix={arXiv},
      primaryClass={cs.CV},
      url={https://arxiv.org/abs/2502.14786}, 
}

@misc{wang2024qwen2vlenhancingvisionlanguagemodels,
      title={Qwen2-VL: Enhancing Vision-Language Model's Perception of the World at Any Resolution}, 
      author={Peng Wang and Shuai Bai and Sinan Tan and Shijie Wang and Zhihao Fan and Jinze Bai and Keqin Chen and Xuejing Liu and Jialin Wang and Wenbin Ge and Yang Fan and Kai Dang and Mengfei Du and Xuancheng Ren and Rui Men and Dayiheng Liu and Chang Zhou and Jingren Zhou and Junyang Lin},
      year={2024},
      eprint={2409.12191},
      archivePrefix={arXiv},
      primaryClass={cs.CV},
      url={https://arxiv.org/abs/2409.12191}, 
}

@misc{beyer2024paligemmaversatile3bvlm,
      title={PaliGemma: A versatile 3B VLM for transfer}, 
      author={Lucas Beyer and Andreas Steiner and André Susano Pinto and Alexander Kolesnikov and Xiao Wang and Daniel Salz and Maxim Neumann and Ibrahim Alabdulmohsin and Michael Tschannen and Emanuele Bugliarello and Thomas Unterthiner and Daniel Keysers and Skanda Koppula and Fangyu Liu and Adam Grycner and Alexey Gritsenko and Neil Houlsby and Manoj Kumar and Keran Rong and Julian Eisenschlos and Rishabh Kabra and Matthias Bauer and Matko Bošnjak and Xi Chen and Matthias Minderer and Paul Voigtlaender and Ioana Bica and Ivana Balazevic and Joan Puigcerver and Pinelopi Papalampidi and Olivier Henaff and Xi Xiong and Radu Soricut and Jeremiah Harmsen and Xiaohua Zhai},
      year={2024},
      eprint={2407.07726},
      archivePrefix={arXiv},
      primaryClass={cs.CV},
      url={https://arxiv.org/abs/2407.07726}, 
}

@misc{li2025ariaopenmultimodalnative,
      title={Aria: An Open Multimodal Native Mixture-of-Experts Model}, 
      author={Dongxu Li and Yudong Liu and Haoning Wu and Yue Wang and Zhiqi Shen and Bowen Qu and Xinyao Niu and Fan Zhou and Chengen Huang and Yanpeng Li and Chongyan Zhu and Xiaoyi Ren and Chao Li and Yifan Ye and Peng Liu and Lihuan Zhang and Hanshu Yan and Guoyin Wang and Bei Chen and Junnan Li},
      year={2025},
      eprint={2410.05993},
      archivePrefix={arXiv},
      primaryClass={cs.CV},
      url={https://arxiv.org/abs/2410.05993}, 
}

@misc{vqvae_oord2018neuraldiscreterepresentationlearning,
      title={Neural Discrete Representation Learning}, 
      author={Aaron van den Oord and Oriol Vinyals and Koray Kavukcuoglu},
      year={2018},
      eprint={1711.00937},
      archivePrefix={arXiv},
      primaryClass={cs.LG},
      url={https://arxiv.org/abs/1711.00937}, 
}

@misc{vqgan_esser2021tamingtransformershighresolutionimage,
      title={Taming Transformers for High-Resolution Image Synthesis}, 
      author={Patrick Esser and Robin Rombach and Björn Ommer},
      year={2021},
      eprint={2012.09841},
      archivePrefix={arXiv},
      primaryClass={cs.CV},
      url={https://arxiv.org/abs/2012.09841}, 
}

@inproceedings{
ma2024at,
title={At Which Training Stage Does Code Data Help {LLM}s Reasoning?},
author={YINGWEI MA and Yue Liu and Yue Yu and Yuanliang Zhang and Yu Jiang and Changjian Wang and Shanshan Li},
booktitle={The Twelfth International Conference on Learning Representations},
year={2024},
url={https://openreview.net/forum?id=KIPJKST4gw}
}

@misc{grad_surgery,
      title={Gradient Surgery for Multi-Task Learning}, 
      author={Tianhe Yu and Saurabh Kumar and Abhishek Gupta and Sergey Levine and Karol Hausman and Chelsea Finn},
      year={2020},
      eprint={2001.06782},
      archivePrefix={arXiv},
      primaryClass={cs.LG},
      url={https://arxiv.org/abs/2001.06782}, 
}

@misc{recon_grad,
      title={Recon: Reducing Conflicting Gradients from the Root for Multi-Task Learning}, 
      author={Guangyuan Shi and Qimai Li and Wenlong Zhang and Jiaxin Chen and Xiao-Ming Wu},
      year={2023},
      eprint={2302.11289},
      archivePrefix={arXiv},
      primaryClass={cs.LG},
      url={https://arxiv.org/abs/2302.11289}, 
}

@misc{t2i_compbench,
      title={T2I-CompBench++: An Enhanced and Comprehensive Benchmark for Compositional Text-to-image Generation}, 
      author={Kaiyi Huang and Chengqi Duan and Kaiyue Sun and Enze Xie and Zhenguo Li and Xihui Liu},
      year={2025},
      eprint={2307.06350},
      archivePrefix={arXiv},
      primaryClass={cs.CV},
      url={https://arxiv.org/abs/2307.06350}, 
}

@misc{mscoco,
      title={Microsoft COCO: Common Objects in Context}, 
      author={Tsung-Yi Lin and Michael Maire and Serge Belongie and Lubomir Bourdev and Ross Girshick and James Hays and Pietro Perona and Deva Ramanan and C. Lawrence Zitnick and Piotr Dollár},
      year={2015},
      eprint={1405.0312},
      archivePrefix={arXiv},
      primaryClass={cs.CV},
      url={https://arxiv.org/abs/1405.0312}, 
}

@misc{unitoken,
      title={UniToken: Harmonizing Multimodal Understanding and Generation through Unified Visual Encoding}, 
      author={Yang Jiao and Haibo Qiu and Zequn Jie and Shaoxiang Chen and Jingjing Chen and Lin Ma and Yu-Gang Jiang},
      year={2025},
      eprint={2504.04423},
      archivePrefix={arXiv},
      primaryClass={cs.CV},
      url={https://arxiv.org/abs/2504.04423}, 
}

@misc{img_edit,
      title={ImgEdit: A Unified Image Editing Dataset and Benchmark}, 
      author={Yang Ye and Xianyi He and Zongjian Li and Bin Lin and Shenghai Yuan and Zhiyuan Yan and Bohan Hou and Li Yuan},
      year={2025},
      eprint={2505.20275},
      archivePrefix={arXiv},
      primaryClass={cs.CV},
      url={https://arxiv.org/abs/2505.20275}, 
}

@misc{ma2023trainingstagedoescode,
      title={At Which Training Stage Does Code Data Help LLMs Reasoning?}, 
      author={Yingwei Ma and Yue Liu and Yue Yu and Yuanliang Zhang and Yu Jiang and Changjian Wang and Shanshan Li},
      year={2023},
      eprint={2309.16298},
      archivePrefix={arXiv},
      primaryClass={cs.CL},
      url={https://arxiv.org/abs/2309.16298}, 
}

@misc{shao2024deepseekmathpushinglimitsmathematical,
      title={DeepSeekMath: Pushing the Limits of Mathematical Reasoning in Open Language Models}, 
      author={Zhihong Shao and Peiyi Wang and Qihao Zhu and Runxin Xu and Junxiao Song and Xiao Bi and Haowei Zhang and Mingchuan Zhang and Y. K. Li and Y. Wu and Daya Guo},
      year={2024},
      eprint={2402.03300},
      archivePrefix={arXiv},
      primaryClass={cs.CL},
      url={https://arxiv.org/abs/2402.03300}, 
}

@article{lin2025toklip,
  title={Toklip: Marry visual tokens to clip for multimodal comprehension and generation},
  author={Lin, Haokun and Wang, Teng and Ge, Yixiao and Ge, Yuying and Lu, Zhichao and Wei, Ying and Zhang, Qingfu and Sun, Zhenan and Shan, Ying},
  journal={arXiv preprint arXiv:2505.05422},
  year={2025},
  url={https://arxiv.org/abs/2505.05422}
}

@article{li2025onecat,
  title={Onecat: Decoder-only auto-regressive model for unified understanding and generation},
  author={Li, Han and Peng, Xinyu and Wang, Yaoming and Peng, Zelin and Chen, Xin and Weng, Rongxiang and Wang, Jingang and Cai, Xunliang and Dai, Wenrui and Xiong, Hongkai},
  journal={arXiv preprint arXiv:2509.03498},
  year={2025},
  url={https://arxiv.org/abs/2509.03498}
}

@article{zhuang2025vargpt,
  title={Vargpt: Unified understanding and generation in a visual autoregressive multimodal large language model},
  author={Zhuang, Xianwei and Xie, Yuxin and Deng, Yufan and Liang, Liming and Ru, Jinghan and Yin, Yuguo and Zou, Yuexian},
  journal={arXiv preprint arXiv:2501.12327},
  year={2025},
  url={https://arxiv.org/abs/2501.12327}
}

@inproceedings{wu2025harmonizing,
  title={Harmonizing visual representations for unified multimodal understanding and generation},
  author={Wu, Size and Zhang, Wenwei and Xu, Lumin and Jin, Sheng and Wu, Zhonghua and Tao, Qingyi and Liu, Wentao and Li, Wei and Loy, Chen Change},
  booktitle={Proceedings of the IEEE/CVF International Conference on Computer Vision},
  pages={17739--17750},
  year={2025},
  url={https://openaccess.thecvf.com/content/ICCV2025/papers/Wu_Harmonizing_Visual_Representations_for_Unified_Multimodal_Understanding_and_Generation_ICCV_2025_paper.pdf}
}

@article{pan2025transfer,
  title={Transfer between modalities with metaqueries},
  author={Pan, Xichen and Shukla, Satya Narayan and Singh, Aashu and Zhao, Zhuokai and Mishra, Shlok Kumar and Wang, Jialiang and Xu, Zhiyang and Chen, Jiuhai and Li, Kunpeng and Juefei-Xu, Felix and others},
  journal={arXiv preprint arXiv:2504.06256},
  year={2025},
  url={https://arxiv.org/abs/2504.06256}
}

@inproceedings{hao2025omnikv,
  title={{OmniKV: Dynamic Context Selection for Efficient Long-Context LLMs}},
  author={Hao, Jitai and Zhu, Yuke and Wang, Tian and Yu, Jun and Xin, Xin and Zheng, Bo and Ren, Zhaochun and Guo, Sheng},
  booktitle={The Thirteenth International Conference on Learning Representations},
  year={2025},
  url={https://openreview.net/forum?id=ulCAPXYXfa}
}

@inproceedings{hao2025token,
  title={{A Token is Worth over 1,000 Tokens: Efficient Knowledge Distillation through Low-Rank Clone}},
  author={Hao, Jitai and Huang, Qiang and Liu, Hao and Xiao, Xinyan and Ren, Zhaochun and Yu, Jun},
  booktitle={The Thirty-ninth Annual Conference on Neural Information Processing Systems},
  year={2025},
  url={https://openreview.net/forum?id=LVDRJE4xQ2}
}

@inproceedings{shenefficient,
  title={Efficient Multimodal Spatial Reasoning via Dynamic and Asymmetric Routing},
  author={Shen, Yixian and Bi, Qi and Wang, Zihan and Yang, Zhiheng and Wang, Changshuo and Zhang, Zhi and Tiwari, Prayag and Pimentel, Andy D and Pathania, Anuj},
  booktitle={The Fourteenth International Conference on Learning Representations},
  year={2026},
  url={https://openreview.net/forum?id=BQASoLmREU}
}

@inproceedings{zhou2025dogr,
  title={Dogr: Towards versatile visual document grounding and referring},
  author={Zhou, Yinan and Chen, Yuxin and Lin, Haokun and Wu, Yichen and Yang, Shuyu and Qi, Zhongang and Ma, Chen and Zhu, Li},
  booktitle={Proceedings of the IEEE/CVF International Conference on Computer Vision},
  pages={3596--3606},
  year={2025},
  url={https://openaccess.thecvf.com/content/ICCV2025/papers/Zhou_DOGR_Towards_Versatile_Visual_Document_Grounding_and_Referring_ICCV_2025_paper.pdf}
}

@article{zhao2026unifying,
  title={Unifying Search and Recommendation in LLMs via Gradient Multi-Subspace Tuning},
  author={Zhao, Jujia and Wang, Zihan and Pan, Shuaiqun and Verberne, Suzan and Ren, Zhaochun},
  journal={arXiv preprint arXiv:2601.09496},
  year={2026},
  url={https://arxiv.org/abs/2601.09496}
}

@inproceedings{wang2025cooperative,
  title={A cooperative multi-agent framework for zero-shot named entity recognition},
  author={Wang, Zihan and Zhao, Ziqi and Lyu, Yougang and Chen, Zhumin and de Rijke, Maarten and Ren, Zhaochun},
  booktitle={Proceedings of the ACM on Web Conference 2025},
  pages={4183--4195},
  year={2025},
  url={https://dl.acm.org/doi/10.1145/3696410.3714923}
}

@inproceedings{wu2025mitigating,
  title={Mitigating hallucinations in large vision-language models via entity-centric multimodal preference optimization},
  author={Wu, Jiulong and Shi, Zhengliang and Wang, Shuaiqiang and Huang, Jizhou and Yin, Dawei and Yan, Lingyong and Cao, Min and Zhang, Min},
  booktitle={Proceedings of the 2025 Conference on Empirical Methods in Natural Language Processing},
  pages={19456--19472},
  year={2025},
  url={https://aclanthology.org/2025.emnlp-main.982.pdf}
}

@article{shi2025deep,
  title={Deep research: A systematic survey},
  author={Shi, Zhengliang and Chen, Yiqun and Li, Haitao and Sun, Weiwei and Ni, Shiyu and Lyu, Yougang and Fan, Run-Ze and Jin, Bowen and Weng, Yixuan and Zhu, Minjun and others},
  journal={arXiv preprint arXiv:2512.02038},
  year={2025},
  url={https://arxiv.org/abs/2512.02038}
}

@article{shi2025iterative,
  title={Iterative self-incentivization empowers large language models as agentic searchers},
  author={Shi, Zhengliang and Yan, Lingyong and Yin, Dawei and Verberne, Suzan and de Rijke, Maarten and Ren, Zhaochun},
  journal={arXiv preprint arXiv:2505.20128},
  year={2025},
  url={https://arxiv.org/abs/2505.20128}
}

@article{hao2026deltakv,
  title={{DeltaKV: Residual-Based KV Cache Compression via Long-Range Similarity}},
  author={Hao, Jitai and Huang, Qiang and Wang, Yaowei and Zhang, Min and Yu, Jun},
  journal={arXiv preprint arXiv:2602.08005},
  year={2026},
  url={https://arxiv.org/abs/2602.08005}
}

@inproceedings{lyumacpo,
  title={MACPO: Weak-to-Strong Alignment via Multi-Agent Contrastive Preference Optimization},
  author={Lyu, Yougang and Yan, Lingyong and Wang, Zihan and Yin, Dawei and Ren, Pengjie and de Rijke, Maarten and Ren, Zhaochun},
  booktitle={The Thirteenth International Conference on Learning Representations},
  year={2025},
  url={https://openreview.net/forum?id=x1Okv4kbVR}
}

@article{lyu2025deepshop,
  title={Deepshop: A benchmark for deep research shopping agents},
  author={Lyu, Yougang and Zhang, Xiaoyu and Yan, Lingyong and de Rijke, Maarten and Ren, Zhaochun and Chen, Xiuying},
  journal={arXiv preprint arXiv:2506.02839},
  year={2025},
  url={https://arxiv.org/abs/2506.02839}
}
\bibliographystyle{iclr2026}

\appendix

\section{Appendix}

\subsection{Other Related Work}
\label{appendix:other_related_work}

\paragraph{Other UMMs}
Recent advancements in Unified Multimodal Models (UMMs) have explored various architectural and methodological innovations. For instance, OneCAT \citep{li2025onecat} proposes a decoder-only autoregressive architecture that utilizes modality-specific Mixture-of-Experts (MoE) structures to achieve efficient multimodal fusion without visual encoders. VARGPT \citep{zhuang2025vargpt} unifies visual understanding and generation through a seamless combination of next-token and next-scale prediction paradigms. To harmonize multi-granularity representations, the Harmon framework \citep{wu2025harmonizing} leverages a shared masked autoregressive encoder. Furthermore, MetaQueries \citep{pan2025transfer} introduces learnable queries to transfer reasoning capabilities from frozen MLLMs to diffusion decoders.

\paragraph{UMMs for Efficient Agent}
The capabilities of vision-language models are continuously evolving, driving their application in specialized vertical domains and complex agentic tasks. For example, the emerging ``Deep Research'' paradigm \citep{shi2025deep}, iterative self-incentivization frameworks \citep{shi2025iterative}, and domain-specific benchmarks like DeepShop \citep{lyu2025deepshop} highlight the critical importance of tool utilization and autonomous multi-step interactions in real-world scenarios. Currently, translating these capabilities into complex tasks often relies on cumbersome multi-agent or multi-model pipelines. Instances of this include cooperative multi-agent systems for entity recognition \citep{wang2025cooperative} and alignment optimization \citep{lyumacpo}, as well as specialized fragmented models for fine-grained document grounding \citep{zhou2025dogr}, search and recommendation \citep{zhao2026unifying}, and preference optimization \citep{wu2025mitigating}. However, these multi-model pipelines inevitably generate extremely long contexts and heavy computational burdens. While these efficiency issues can be partially alleviated by inference acceleration and efficient training methods like OmniKV~\citep{hao2025omnikv}, SnapKV~\citep{li_snapkv_2024}, and Low-Rank Clone \citep{hao2025token}, relying on fragmented systems lacks architectural elegance~\citep{shenefficient,hao2026deltakv}. In contrast, UMMs offer a more concise and efficient alternative, enabling a single unified model to seamlessly handle diverse reasoning, tool-calling, and multimodal tasks without the cumbersome overhead of complex multi-agent interactions.

\subsection{Prompts of Image Generation}
\label{appendix:prompts_gen_img}

Table~\ref{tab:prompts_of_img_gen} lists the prompts corresponding to the generated images shown in Figure~\ref{fig:gen_case}. 
The prompts are presented in the same order as the images: left to right, top to bottom. These examples highlight the diversity of tasks, ranging from descriptive captions to creative scene generation.

\begin{table}[ht]
\centering
\renewcommand{\arraystretch}{1.1}
\caption{Prompts used for the image generation examples shown in Figure~\ref{fig:gen_case}.}
\label{tab:prompts_of_img_gen}
\vspace{0.5em}
\begin{tabular}{l p{0.9\textwidth}}
  \toprule
  \rowcolor[HTML]{DDEBFF} \textbf{No.} & \textbf{Prompt} \\
  \midrule
  1 & A gigantic library floats above the clouds, its appearance resembling a suspended castle. Every book emits a faint glow and drifts through the air with the gentle breeze. \\
  2 & A highly realistic close-up photo featuring a beautiful 35-year-old red-haired woman, writing in her diary on her balcony. She is dressed in warm yet stylish clothing. \\
  3 & A happy snowman. \\
  4 & A woman and her little lion taking a selfie on the grassland. \\
  5 & A beautiful owl with sleek feathers and lively eyes, its round head adorned with two furry ears. The elegant pattern is formed by the interweaving of snow-white down and deep brown flight feathers, making it appear both stunning and endearing. \\
  6 & A clearing in a deep, mysterious forest, with a mirror-like pond at its center, the water reflecting a night sky filled with the Milky Way. \\
  7 & A handsome 24-year-old boy stands in the center, with a sky-colored background. He is wearing glasses, and the art style is very detailed, in anime style. \\ 
  8 & A realistic photo of an elephant walking on the ocean floor. \\
  9 & An elegant and charming lady whose hair is entirely made up of blooming flowers, resembling a masterpiece of nature. The flowers are of various types, possibly including delicate roses, fresh daisies, vibrant sunflowers, or other colorful blossoms. \\
  10 & A magnificent landscape photo depicting the northern lights dancing above the snow-capped mountain ranges in Iceland. \\
  \bottomrule
\end{tabular}
\end{table}
\setlength{\textfloatsep}{1.0em}

\subsection{Baseline Implementation Details}
\label{appendix:baseline_impl}

To ensure fair comparisons, we adapt baseline methods to the VQ+AR setting used in our study.
%%%
For Mixture-of-Transformers (MoT)~\citep{deng_emerging_2025, shi_lmfusion_2025, liao_mogao_2025}, the duplicated transformer is originally designed for image generation through diffusion. 
To remove the influence of diffusion and isolate architectural effects, we reconfigure the duplicated transformer to operate directly on image tokens.
In this setup, the \texttt{qkv} sequences from the two transformers are concatenated within the attention module, allowing the model to incorporate visual information for both understanding and generation tasks. 
%%%
As a result, the MoT results reported in this paper reflect its effectiveness strictly within the VQ+AR paradigm, eliminating confounding factors introduced by diffusion-based processes.

\subsection{More Evaluation Results on Image Generation Benchmark}
\label{appendix:more_results_on_img_gen}

\revise{We conducted tests on the T2I-CompBench~\citep{t2i_compbench} and MSCOCO~\citep{mscoco}. Part of the results were excerpted from UniToken~\citep{unitoken}. As shown in Table~\ref{tab:more_gen_img_results}, Uni-X surpassed the recent strong autoregressive models EMU3 and Liquid in the newly added image generation benchmark. Uni-X also achieved better results than UniToken, which includes semantic information.}

\begin{table}[t]
\centering
\small
\renewcommand{\arraystretch}{1.1}
\caption{The T2I-CompBench and MSCOCO performance of Uni-X. $^\heartsuit$ indicates that it has been trained on more image-text data.}
\label{tab:more_gen_img_results}
\vspace{0.5em}
\resizebox{\textwidth}{!}{
\begin{tabular}{llrrrrrr}
    \toprule
    \rowcolor[HTML]{DDEBFF} \textbf{Model} & \textbf{\# Params.} & \textbf{T2I-Color} & \textbf{T2I-Shape} & \textbf{T2I-Texture} & \textbf{T2I-Avg. $\uparrow$} & \textbf{MSCOCO CLIP-T} \\
    \midrule
    SDXL & 3.5B & 63.7 & 54.1 & 56.4 & 58.1 & -- \\
    Janus & 1.3B & 75.5 & 47.7 & 62.1 & 61.8 & -- \\
    Liquid & 7B & 71.5 & 52.3 & 65.1 & 63.0 & 30.7 \\
    EMU3 & 8B & 61.1 & 47.3 & 61.9 & 56.8 & 31.3 \\
    UniToken & 7B & 71.2 & 51.8 & 66.7 & 63.2 & -- \\
    \midrule
    \rowcolor[HTML]{FFF2CC} \textbf{Uni-X}$^\heartsuit$ & \textbf{3B / 4.5B} & \textbf{76.5} & \textbf{56.3} & \textbf{67.1} & \textbf{66.6} & \textbf{31.8} \\
    \bottomrule
\end{tabular}}
\end{table}

\subsection{Gradient Conflict Analysis}
\label{appendix:more_grad_conflict}

\revise{\paragraph{Analysis on Mainstream Models.} We further demonstrate the effectiveness of the current gradient conflict metric through experiments. We conduct a quantitative analysis on mainstream models such as Qwen and Llama, as shown in Table~\ref{tab:gradient_conflict_mainstream}. For each dataset, we utilized a total of 2M tokens (accumulated over 60 batches) to compute gradients, to ensure minimal gradient noise.

All models in Table~\ref{tab:gradient_conflict_mainstream} exhibit a consistent pattern: the gradient conflict between Code vs. Math is strictly lower than for both Code vs. Wiki and Math vs. Wiki. It is well-established in LLM pre-training that Code and Math tasks often mutually enhance each other~\citep{shao2024deepseekmathpushinglimitsmathematical, ma2023trainingstagedoescode}. This phenomenon is precisely reflected in our gradient conflict analysis.

The relatively high gradient similarity (low conflict) between these two tasks implies that improvements in Code performance can drive improvements in Math performance. We further verified this in Table~\ref{tab:gradient_conflict_mainstream_impact}. Qwen2.5-Coder-3B, which was fine-tuned from Qwen2.5-3B to specifically enhance coding capabilities, simultaneously achieved a substantial improvement in Math performance. This validates our hypothesis that lower gradient conflict correlates with positive transfer between modalities/domains.}

\begin{table}[t]
\centering
\small
\renewcommand{\arraystretch}{1.1}
\caption{Average gradient conflict between different domain data. Higher values indicate a higher degree of conflict.}
\label{tab:gradient_conflict_mainstream}
\vspace{0.5em}
\begin{tabular}{lrrr}
  \toprule
  \rowcolor[HTML]{DDEBFF}  \textbf{Model} & \textbf{Code vs. Math} & \textbf{Code vs. Wiki} & \textbf{Math vs. Wiki} \\
  \midrule
  Qwen2.5-1.5B     & 0.158         & 0.330         & 0.262         \\
  Qwen2.5-3B       & 0.130         & 0.382         & 0.294         \\
  Qwen2.5-Coder-3B & 0.182         & 0.317         & 0.275         \\
  Qwen2.5-7B       & 0.153         & 0.263         & 0.240         \\
  Llama3.2-3B      & 0.297         & 0.351         & 0.360         \\
  \bottomrule
\end{tabular}
\end{table}

\begin{table}[t]
\centering
\small
\renewcommand{\arraystretch}{1.1}
\caption{Domain performance of Qwen2.5-3B and Qwen2.5-Coder-3B under zero-shot settings.}
\label{tab:gradient_conflict_mainstream_impact}
\vspace{0.5em}
\begin{tabular}{lrrr}
  \toprule
  \rowcolor[HTML]{DDEBFF} \textbf{Model} & \textbf{HumanEval (Code)} & \textbf{GSM8K (Math)} & \textbf{MMLU (Wiki)} \\
  \midrule
  Qwen2.5-3B       & 39.0             & 6.0          & 65.0        \\
  Qwen2.5-Coder-3B & 45.7             & 26.1         & 60.8        \\
  \bottomrule
\end{tabular}
\end{table}

\paragraph{Analysis on Other Modules.} In Section~\ref{sect:method:obser} of the main text, we analyzed gradient conflicts in the down-projection weights of the Feed-Forward Network (FFN). To develop a more complete picture and confirm that this issue is not confined to a single component, we extend our analysis to additional modules of the transformer.
In particular, we examine gradient conflicts in the output projection weights (\texttt{O\_PROJ}) and value projection weights (\texttt{V\_PROJ}) of the self-attention mechanism, both of which play critical roles in multimodal representation learning.

\begin{figure}[ht]
    \centering % 整体居中
    \begin{minipage}{0.47\textwidth}
        \centering % 图片在minipage内居中
        \includegraphics[width=0.99\textwidth]{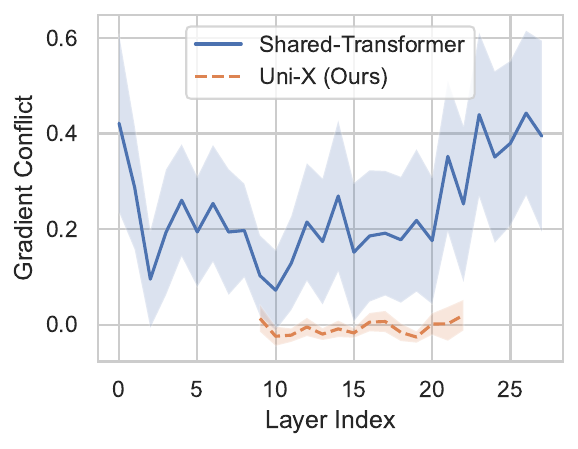}
        \vspace{-2.0em}
        \caption{An analysis of gradient conflict in attention of out projection weights.}
        \label{fig:grad_conflict_out}
        \vspace{-0.25em}
    \end{minipage}
    \hfill % 两个minipage之间添加水平填充（空格）
    \begin{minipage}{0.47\textwidth}
        \centering % 图片在minipage内居中
        \includegraphics[width=0.99\textwidth]{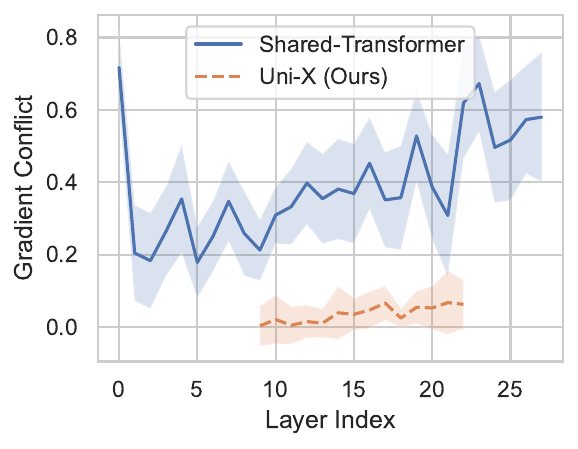}
        \vspace{-2.0em}
        \caption{An analysis of gradient conflict in attention of value projection weights.}
        \label{fig:grad_conflict_out_key}
        \vspace{-0.25em}
    \end{minipage}
    % \caption{两张图片的总标题（可选）} % 可选的总标题
    % \label{fig:both-images} % 总标签（可选）
\end{figure}

Using the same methodology for conflict measurement, Figures~\ref{fig:grad_conflict_out} and \ref{fig:grad_conflict_out_key} reveal a consistent trend with that observed in the FFN layers.
%%%
The modality-shared transformer exhibits severe gradient conflicts in the shallow and deep layers of both the attention output and value projection weights, with only partial alleviation in the middle layers. 
%%%
In contrast, Uni-X effectively addresses these issues: (i) modality-specific layers at both ends prevent conflicts in low-level processing and output stages, and (ii) the shared middle block further reduces residual conflicts by leveraging semantic alignment.

These results strengthen our hypothesis that gradient conflict stems from the intrinsic statistical mismatch between vision and text, and they demonstrate that Uni-X's two-end-separated, middle-shared design offers a robust and generalizable solution across multiple transformer components.

\revise{\subsection{Inference Efficiency}
\label{appendix:infer_eff}

We have conducted a comprehensive evaluation of inference efficiency on an H800 PCIe (350W) GPU. As shown in Table~\ref{tab:inference_throughput_comparison}, Uni-X demonstrates superior throughput compared to standard autoregressive baselines.
%%%
Uni-X achieves high throughput (910.2 tokens/s) even compared to the original Qwen2.5-3B (975.2 tokens/s), despite the architectural changes and a higher number of parameters (4.5B vs 3B). This efficiency gain stems from the computational complexity of the attention mechanism in the separated layers.

Theoretically, the computational cost of the Uni-X architecture is lower, and the current inference speed still has a slight gap because the current code has not been fully optimized. In the separated layers, a sequence of length $n$ is effectively partitioned into vision tokens of length $a$ and text tokens of length $b$ (where $a+b=n$). Since the self-attention complexity is $O(n^2)$, and the separated layers enforce strict modality isolation, the complexity reduces to proportional to $a^2 + b^2$. Since $a^2 + b^2 < (a+b)^2=n^2$, the computational cost for attention in these specific layers is strictly lower than in a fully shared transformer, leading to the observed speedup.}

\begin{table}[t]
\centering
\small
\renewcommand{\arraystretch}{1.1}
\caption{Inference throughput comparison. Settings: batch size 48, input length $\approx$1,200 tokens, outputting one image.}
\label{tab:inference_throughput_comparison}
\vspace{0.5em}
\begin{tabular}{llrr}
  \toprule
  \rowcolor[HTML]{DDEBFF} \textbf{Model} & \textbf{\# Params.} & \textbf{Tokens/s} & \textbf{Images/min} \\
  \midrule
  % 数据部分（示例）
  Shared Transformer (Qwen2.5-3B) & 3B & 975.2 & - \\
  Liquid & 7B & 182.0 & 10.6 \\
  EMU3 & 8B & 199.0 & 2.9 \\
  \midrule
  \rowcolor[HTML]{FFF2CC} \textbf{Uni-X} & 3B / 4.5B & 910.2 & \textbf{53.3} \\
  \bottomrule
\end{tabular}
\end{table}

% \begin{table}[t]
% \centering
% \small
% \renewcommand{\arraystretch}{1.1}
% \caption{\revise{Inference throughput comparison. Settings: batch size 48, input length $\approx$1,200 tokens, outputting one image.}}
% \label{tab:inference_throughput_comparison}
% \vspace{0.5em}
% \begin{tabular}{llcc} % 注意后两列改为居中(c)或右对齐(r)均可
%     \toprule
%     % 第一行表头：合并 Throughput
%     \multirow{2.5}{*}{\cellcolor[HTML]{DDEBFF}\textbf{Model}} & \multirow{2.5}{*}{\cellcolor[HTML]{DDEBFF}\textbf{\# Params.}} & \multicolumn{2}{c}{\cellcolor[HTML]{DDEBFF}\textbf{Throughput $\uparrow$}} \\    
%     % 分割线：仅在第3到第4列下方划线，(lr)表示左右留白，避免线条相连
%     \cmidrule(lr){3-4}
%     % 第二行表头：具体的单位
%     \rowcolor[HTML]{DDEBFF} & & \textbf{Tokens/s} & \textbf{Images/min} \\
%     \midrule
%     % 数据部分（示例）
%     Shared Transformer (Qwen2.5-3B) & 3B & 975.2 & - \\
%     Liquid & 7B & 182.0 & 10.6 \\
%     EMU3 & 8B & 199.0 & 2.9 \\
%     \midrule
%     \rowcolor[HTML]{FFF2CC} \textbf{Uni-X} & 3B / 4.5B & 910.2 & \textbf{53.3} \\
%     \bottomrule
% \end{tabular}
% \end{table}

\revise{\subsection{Ablation Study on Ratio between Text and Vision.}
\label{appendix:tv_ratio}

We conducted experiments maintaining the same hyperparameters and training volume as in Table~\ref{tab:uni_x_configs}, and the results are shown in Table~\ref{tab:tv_ratio}. We continued to use Qwen2.5-1.5B with a total of 28 layers as the base model. The number of vision layers directly affects the performance related to image understanding and generation. Surprisingly, reducing the number of vision layers also decreases pure text performance. This may be because the shared layers in the middle have to process more low-level vision information, thereby leading to a decline in pure text capability. This experimental result also proves the effectiveness of our proposed architecture from another perspective.}

\begin{table}[t]
\centering
\small
\renewcommand{\arraystretch}{1.2}
\caption{\revise{Ratio between t-layers and v-layers within the separated layers.}}
\label{tab:tv_ratio}
\vspace{0.5em}
\begin{tabular}{lrrrr}
\toprule
\rowcolor[HTML]{DDEBFF} \textbf{Configuration} & \textbf{MMLU} & \textbf{GenEval} & \textbf{MMB}  & \textbf{Avg. $\uparrow$} \\
\midrule
14:8                    & 48.2 & 37.8    & 26.1 & 37.4            \\
14:14                   & 49.6 & 41.3    & 29.4 & 40.1            \\
14:20                   & 50.1 & 42.6    & 31.0 & 41.2            \\
\bottomrule
\end{tabular}
\end{table}

\subsection{Use of Large Language Models}
\label{appendix:llm-usage}

Large Language Models (LLMs) were used solely as writing aids during manuscript preparation. 
Their role was limited to language polishing, improving grammar, clarity, and readability, without influencing the conceptual design, experimental methodology, or analytical findings. 
All research ideas, model designs, and experimental results are the original contributions of the authors.

\end{document}